\pdfoutput=1

\documentclass[tikz,border=12pt]{report}

\usepackage[linguistics]{forest}

\usepackage{nert}

\makeatletter
\@removefromreset{ExNo}{chapter}
\makeatother

\usepackage{times}
\usepackage{latexsym}

\usepackage[utf8]{inputenc}

\usepackage{times}
\usepackage{xcolor}
\usepackage[T1]{fontenc}
\usepackage[utf8]{inputenc}
\usepackage{rotating}
\usepackage{needspace}
\usepackage{graphicx}
\usepackage{hyperref}
\usepackage{multicol}
\makesavenoteenv{tabular} 
\makesavenoteenv{tabbing} 

\newcommand{\func}[1]{\textsf{#1}}
\newcommand{\mention}[1]{\textit{#1}}

\title{CGELBank Annotation Manual v1.2}
\author{Brett Reynolds, Nathan Schneider, and Aryaman Arora}
\date{}

\newcommand{\Node}[2]{\small\textsf{#1:}\\{#2}}
\newcommand{\Head}[1]{\Node{Head}{#1}}
\newcommand{\Subj}[1]{\Node{Subj}{#1}}
\newcommand{\Comp}[1]{\Node{Comp}{#1}}
\newcommand{\Mod}[1]{\Node{Mod}{#1}}
\newcommand{\Det}[1]{\Node{Det}{#1}}
\newcommand{\PredComp}[1]{\Node{PredComp}{#1}}

\newcommand{\Mk}[1]{\Node{Marker}{#1}}
\newcommand{\Obj}[1]{\Node{Obj}{#1}}
\newcommand{\Sup}[1]{\Node{Supplement}{#1}}

\newcommand{\Vaux}[0]{V\textsubscript{aux}\xspace}
\newcommand{\Clauserel}[0]{Clause\textsubscript{rel}\xspace}

\begin{document}

\maketitle
\tableofcontents
\chapter{Introduction}

CGELBank \citep{cgelbank-law} is a treebank and associated tools based on a syntactic formalism for English derived from the \emph{Cambridge Grammar of the English Language} \citep[CGEL;][]{cgel}.\footnote{Page references in this document are to pages in CGEL unless otherwise noted.} It is hosted on GitHub at \url{https://github.com/nert-nlp/cgel}.
This document lays out the particularities of the CGELBank annotation scheme.

\bigskip

As CGELBank is based on CGEL, CGEL itself should be the go-to resource for answers to most questions about the framework like ``what is a determiner'' or ``what is the structure of the pseudo-cleft construction?'' As much as possible, CGELBank follows the analysis set out in CGEL (and its companion textbook, \emph{A Student's Introduction to English Grammar, 2nd edition} \citep{sieg2}). But that doesn't mean 100\% correspondence. The purpose of this document, then, is to clarify the approach and highlight cases of non-correspondence. We see four general reasons why CGELBank and CGEL may not correspond precisely.
\begin{enumerate}
    \item We believe CGEL has not been sufficiently explicit or exhaustive.
    \item We believe CGEL has made an error.
    \item We wish to represent something in a simpler or clearer way, even though the first two reasons do not apply.
    \item We have made an error (please, let us know).
\end{enumerate}

The most obvious case of being inexhaustive is in the lexicon. CGEL does not and could not list the category of every English word, nor would that be useful, as most dictionaries do a mostly fine job of identifying nouns and verbs. When it come to the closed classes, though, one might wonder whether \mention{certain}, for instance, is always an adjective or whether it is sometimes a determinative. We attempt to provide exhaustive lists of pronouns, determinatives, prepositions, subordinators, and coordinators.

Another example of an ambiguity in CGEL, this one pertaining to syntax, is whether a relative pronoun in a relative clause like \textit{who ate there} is a \func{prenucleus} followed by a \func{subject} gap or whether it is an in-situ subject. (We take the position that all relative clauses have gaps; see \cref{sec:relatives}.)
Where possible, we turn to subsequent publications by CGEL authors \citep{payne-07,payne-10,payne-13,pullum-13} to resolve ambiguities.

When it comes to errors, we believe there are only a very few in CGEL, and even those we are uncertain about. We believe, though, that the treatment of coordinates as unheaded, is unmotivated for reasons we explain in chapter \ref{ch:coord}. We also believe the analyzing complex numerals as lexemes is an error, for reasons explained in \citet{reynolds2025}.

An example (perhaps the only one) of a different representation -- rather than a different analysis -- relates to the way we portray \func{supplements} in tree structure, a topic also discussed in \cref{sec:Supplements}.


\section{New in v1.2}
\begin{itemize}
    \item new analysis of complex numerals (\cref{sec:complex-num})
    \item revise policy on enumerated proper names to better align with CGEL (\cref{sec:derived})
    \item new section on morphological information present in the data (\cref{sec:xpos})
    \item new section on predicative complements (\cref{sec:predcomp})
    \item clarify the section on punctuation encoding and change the conventions for which lexical node the punctuation token attaches to (\cref{sec:punc-enc})
    \item recognize ternary branching in double-complement PPs: \cref{ex:dblcomppp} in \cref{sec:beyondbinary}
    \item clarify that subjectless complements are clauses not VPs: \cref{ex:NoSubjs} in \cref{sec:subjinfpart}
    \item clarify how to determine the level of coordination (\cref{sec:basiccoord})
    \item clarify NP structure where coordinated heads share a post-head dependent (\cref{sec:shared-post-head})
    \item \mention{so~as} has been removed from the list of complex lexical items (\cref{sec:complex}; see \cref{fn:so-as}), while \mention{so~long~as} has been added
\end{itemize}

\section{New in v1.1}

\begin{itemize}
\item add PP\textsubscript{strand} subcategory (\cref{sec:prepstrand}) and mention of prepositional passive (\cref{sec:preppass})
\item add subsection for fused \func{Marker-Head} function (\cref{sec:marker-head})
\item add hollow purpose \mention{for}-clause example \cref{ex:HollowForPurpose}
\item elaborate elliptical stranding (\cref{sec:ellipstrand})
\item relax ``A gap must have at least one non-gap sister in a function other than \func{supplement}'' from a hard constraint to a tendency, noting a counterexample \cref{ex:doublegap}
\item minor corrections and clarifications
\end{itemize}


\chapter{The lexicon, categories, and functions}\label{ch:lex}
\section{Introduction}
Each lexeme is assigned a lexical category (e.g., noun, verb, adjective). These categories typically project phrases; for example an adjective projects and adjective phrase. The relations between phrases are labeled with functions (e.g., head, object, determiner; \cref{ch:branching} discusses how this is realized in the trees\footnote{In this document, functions appear in a non-serif font like this: \func{Head}.}).

\section{Categories}
CGEL and CGELBank have the following lexical categories and phrasal projections:

\begin{tabbing}
    Noun (N, N\textsubscript{pro}) \hspace{3 em}\=$\rightarrow$ Nominal (Nom)\hspace{2 em} \=$\rightarrow$ Noun phrase (NP)\\
    Verb (V, V\textsubscript{aux}) \>$\rightarrow$ Verb phrase (VP) \>$\rightarrow$ Clause (incl.~Clause\textsubscript{rel}) \\
    Adjective (Adj) \>\>$\rightarrow$ Adjective phrase (AdjP)\\
    Adverb (Adv) \>\>$\rightarrow$ Adverb phrase (AdvP)\\
    Preposition (P) \>\>$\rightarrow$ Preposition phrase (PP, PP\textsubscript{strand})\\
    Determinative (D) \>\>$\rightarrow$ Determinative phrase (DP)\\
    Interjection (Int) \>\>$\rightarrow$ Interjection phrase (IntP)\footnote{See fn.~70 on p.~1361.}\\
    Subordinator (Sdr)\\
    Coordinator
    \label{tab:Cats}
\end{tabbing}
Non-headed non-lexical categories (including Coordination and GAP) are presented in \cref{sec:non-head}.

\subsection{Lexical Projection Principle}\label{sec:lexprojection}

Outside of morphologically derived expressions (\cref{sec:derived}), and excepting coordinators and subordinators, \textbf{a lexical node almost always projects a phrase of the corresponding category} shown in the table above. Thus, every N must serve as head within a Nom; every V must head a VP; every Adj must head an AdjP; and so forth.\footnote{The lexical head may be sister to a complement or modifier, if there is one (\cref{ch:branching}).}

The one exception is that subject-auxiliary inversion (\cref{sec:sai}) targets auxiliaries specifically (rather than the VP they would project in normal position), so if the constituent in \func{Prenucleus} function consists of a single unmodified \Vaux, it will not project a VP there.

\subsection{Two-leveled phrases}
As can be seen in the table above, nouns and verbs project two distinct levels of phrase structure. Nouns function as \func{heads} of Noms, which typically function as \func{heads} of NPs but which may also function as modifiers in a Nom, as in \mention{a \underline{multiple choice} question}.

Similarly, verbs function as \func{heads} of VPs, and VPs typically function as \func{heads} of Clauses, but they may also function as modifiers in a Nom, for instance \mention{the \underline{quickly flowing} water}.

\subsection{Non-headed categories}\label{sec:non-head}
\begin{enumerate}
    \item Coordination\footnote{The categories ``coordinator'' and ``coordination'', along with the function \func{coordinate} are always written out in full to limit confusion.}
    \item GAP (gap terminals are represented with an en-dash `--')
    \item Nonce categories are written with the categories of the child constituents joined by the + operator (e.g., NP + PP for \mention{the children in tow}).
\end{enumerate}

\subsection{Subcategories}


Auxiliary verbs bear the label V\textsubscript{aux}, pronouns bear the label N\textsubscript{pron}, and relative clauses bear the label Clause\textsubscript{rel}.\footnote{Other clause types may be added in the future (\cref{sec:future}).}

\label{sec:prepstrand}
PPs exhibiting preposition stranding have their own subcategory: PP\textsubscript{strand}. This serves to disambiguate cases like \mention{The horse was walked} [$_{\text{PP\textsubscript{*}}}$ \mention{around}]. In the stranded reading, the horse is being avoided by walkers; the clause is a prepositional passive (\cref{sec:preppass}). In the non-stranded reading, \mention{around} is an intransitive preposition and somebody was leading the horse here and there.\footnote{Elliptical stranding (\cref{sec:ellipstrand}), in which an auxiliary or the subordinator \mention{to} appears without a complement, is not subject to the same kind of ambiguity with an intransitive reading, and thus elliptical strandings are not specially labeled in the tree.}

%
%
%

\subsubsection{Not clauses}
The following sentences are main clauses in CGEL (p.~944) but not in CGELBank:

\begin{enumerate}
    \item Clauses with subordinate form (e.g., \mention{That it were true!}). These are subordinate clauses in CGELBank.
    \item Conditional fragments (e.g., \mention{If only I could!}). These are PPs in CGELBank.
    \item Verbless directives (e.g., \mention{Out of my way!} or \mention{This way!}).  These are XPs in CGELBank.\footnote{X is a variable for a lexical category, so an XP is a phrase ultimately headed by an X.}
    \item Parallel structures (e.g., \mention{The sooner, the better!}). These are nonce XP + XP in CGELBank.)
\end{enumerate}

Subordinate verbless clauses such as \mention{With \underline{the kids in tow}, he headed out}, are treated as nonce constituents (here NP + PP).\footnote{\mention{The kids in tow} may be a clause semantically, but a syntactic clause in CGEL is a projection of the VP. In a footnote on p.~1286, CGEL says, that "the ultimate head of \mention{hat in hand} is \mention{in}\ldots, with \mention{hand} an internal complement (\mention{in hand} constituting the predicate) and \mention{hat} an external complement (more specifically, the subject).” This is then a kind of 3rd layer on the PP analogous to the NP over the Nom or the Clause over the VP}


\section{Morphological information}\label{sec:xpos}

A limited amount of morphological information is included in CGELBank data (though not visualized in the trees in this document), namely:

\begin{itemize}
    \item \textbf{Lemma}: the lexical lemma, following English treebanking conventions in the Universal Dependencies project,\footnote{\citet{de_marneffe-21}; \url{https://universaldependencies.org/}} is provided explicitly where it differs from the surface form
    \item \textbf{Correct spelling}, if the surface form is a misspelling
    \item \textbf{Morphologically-refined part of speech (``XPOS'')} for cardinal numbers (\texttt{CD}), list item markers (\texttt{LS}), and verbs, following Penn Treebank and Universal Dependencies notation. The verb XPOS values are explained in the table below:\footnote{In CGEL, some of these verbs are taken to have negative forms as well, whereas in PTB the negative clitic is segmented and thus not reflected in the verb's XPOS. The labeling of forms of \mention{be} above is slightly simplified; see CGEL p.~75 for full terminology including irrealis \mention{were}.}
\end{itemize}

\begin{table}[h]\centering
\begin{tabular}{lllp{4em}p{10em}}
\toprule
    \textbf{XPOS} & \multicolumn{2}{c}{\textbf{CGEL terminology}} & \textbf{Lexical Verb (V)} & \textbf{Auxiliary Verb (\Vaux)} \\ \midrule
    & \textsc{primary forms} \\
    \midrule
    \texttt{VBP} & plain present tense & & e.g. \mention{eat} & \mention{am}, \mention{are}, \mention{have}, \mention{do} \\
    \texttt{VBZ} & 3rd sg present tense & & \mention{eats} & \mention{is}, \mention{has}, \mention{does} \\  \cmidrule{3-3}
    \multirow{2}{1em}{\phantom{xxx} \phantom{xxx} \texttt{MD}} & present tense & \multirow{2}{2.5em}{\centering \phantom{xxxxx} modal aux} & --- & \mention{can}, \mention{may}, \mention{will}, \mention{shall}, \mention{must}, \mention{ought}, \mention{need}, \mention{dare} \\ \cmidrule{2-2}
     & \multirow{2}{3em}{\phantom{xxxxx} preterite} &  & --- & \mention{could}, \mention{might}, \mention{would}, \mention{should} \\  \cmidrule{3-3}
    \texttt{VBD} & &  & \mention{ate} & \mention{was}, \mention{were}, \mention{had}, \mention{did} \\ \midrule
    & \textsc{secondary forms}  \\ \midrule
    \texttt{VB} & plain form & & \mention{eat} & \mention{be}, \mention{have}, \mention{do}  \\
    \texttt{VBG} & gerund-participial & & \mention{eating} & \mention{being}, \mention{having}, \mention{doing} \\
    \texttt{VBN} & past participle & & \mention{eaten} & \mention{been}, \mention{had}, \mention{done} \\ \bottomrule
\end{tabular}
\end{table}

Note that the categorizations are closely aligned between XPOS and CGEL, though the closed class of modal auxiliaries is separated as a top-level category in XPOS only, and their tense is not indicated in XPOS.

Refer to \Cref{ch:format} for format details of how morphological information is encoded in the data.

\section{Lexemes}

Where in doubt about the category of a given lexeme, consult the the \href{https://simple.wiktionary.org}{Simple English Wiktionary}. Note that determinatives are called ``determiners'' there.

\subsection{Small categories}
A mostly exhaustive list of each of the following categories is provided at the Simple English Wiktionary. Follow the links.
\begin{enumerate}
    \item \href{https://simple.wiktionary.org/wiki/Category:Prepositions}{Prepositions}
    \item \href{https://simple.wiktionary.org/wiki/Category:Coordinators}{Coordinators}
    \item \href{https://simple.wiktionary.org/wiki/Category:Subordinators}{Subordinators}
    \item \href{https://simple.wiktionary.org/wiki/Category:Determiners}{Determinatives} (called ``determiners'' on the Wiktionary)
\end{enumerate}

\subsection{Complex lexical items (written with a space)}\label{sec:complex}
CGELBank includes a small number of complex lexemes listed in \cref{ex:MWI}, as follows: (a) determinatives, (b) ordinal adjectives, (c) fractional nouns, (d) prepositions, (e) subordinators, and (f) coordinators.

\ex. \label{ex:MWI}
    \a. \a. \mention{a certain}\hspace{2 em}\mention{a few}\hspace{2 em}\mention{a great many}\hspace{2 em}\mention{a little}\hspace{2 em}\mention{many a}\hspace{2 em}\mention{no one}
        \b. \mention{twenty one} {\dots} \mention{twenty nine}  \hspace{2 em}\dots\hspace{2 em}\mention{ninety one} {\dots} \mention{ninety nine}
        \z.
    \b. \mention{twenty first} {\dots} \mention{twenty ninth}  \hspace{2 em}\dots\hspace{2 em}\mention{ninety first} {\dots} \mention{ninety ninth}
    \c. \mention{twenty first} {\dots} \mention{twenty ninth} \hspace{2 em}\dots\hspace{2 em}\mention{ninety first} {\dots} \mention{ninety ninth}
    \c. \a.\label{ex:compound-prep} 
        \mention{as for}
        \hspace{2 em}\mention{as from}
        \hspace{2 em}\mention{as if}
        \hspace{2 em}\mention{as of}
        \hspace{2 em}\mention{as per}
        \hspace{2 em}\mention{as though}
        \hspace{2 em}\mention{as~to}
        \hspace{2 em}\mention{in case}
        \hspace{2 em}\mention{in charge}
        \hspace{2 em}\mention{in front}
        \hspace{2 em}\mention{in order}
        \hspace{2 em}\mention{in spite}
        \hspace{2 em}\mention{in view}
        \hspace{2 em}\mention{no matter}
        \hspace{2 em}\mention{on board}
        \hspace{2 em}\mention{on~purpose}
        \hspace{2 em}\mention{on~to}
        \hspace{2 em}\mention{on~top}
        \hspace{2 em}\mention{à la}
        \b. 
        \mention{as long as} (`provided') 
        \hspace{1 em}\mention{as soon as} (`once')
        \hspace{1 em}\mention{so long as} (`provided') 
        \z.
    \c.   \mention{whether or not} 
        \hspace{2 em}\mention{whether or no}
    \d.    \mention{as well as}
        \hspace{2 em}\mention{rather than}
    \z.
\z.

These items exhibit a high degree of grammaticalization, preventing their parts from being analyzed as syntactically separate units.\footnote{\label{fn:so-as}Despite a suggestion on p.~727 that \mention{so~as} is a preposition, p.~1264 gives a more explicit analysis, arguing that preposition \mention{so} licenses \mention{as}-PP complements. We have therefore removed \mention{so~as} from \cref{ex:compound-prep}.}

We take items such as \mention{We plan to \underline{flight test} and operate it} to be compounds (pp.~1644--1661), not complex lexical items. They should be rewritten with a hyphen.

\subsection{Morphologically derived complex expressions}\label{sec:derived}

Other expressions written as multiple words are treated as derived from a process of (productive) compounding that is more morphological than syntactic.

\subsubsection{Flat nouns and determinatives}\label{sec:flat}
Certain expressions comprised of nouns but lacking ordinary headed NP structure are considered ``flat'' expressions \citep[borrowing the terminology of the Universal Dependencies project;][]{de_marneffe-21}:

\needspace{5cm}

\ex. \label{ex:Flat} 
    \begin{forest}
    where n children=0{
        font=\itshape, 			
        tier=word          			
      }{
      },
      [NP
          [\Head{Nom}
            [\Head{N}
                [\Node{Flat}{N}[Osama]]
                [\Node{Flat}{N}[bin]]
                [\Node{Flat}{N}[Laden]]
            ]
          ]
        ]
    \end{forest}

Multiword proper names (especially personal names: \cref{ex:Flat}), dates, and terminology (\mention{carbon dioxide}) often fall into this category.

Certain other kinds of specialized name patterns nevertheless permit one part to be identified as a modifier (pp.~517--518). Enumerated nouns like \mention{page~27} or \mention{building~J} we treat as containing an entity type as pre-head modifier and an enumerator as head. (This is consistent with CGEL's example of \mention{Ward~17}, p.~518.)

Proper names derived from ordinary NPs (e.g., \mention{Yanhee Hospital}, \mention{No Time to Die}) receive ordinary NP analyses.


\subsubsection{Zero-derived NPs}

Proper names may take the form of a phrase other than an NP---for example, a title of a book that takes the form of a Clause, VP, or PP (but is treated as an NP with respect to the rest of the sentence).
In such cases, we show the internal syntactic structure of the source phrase at the bottom of the tree, and then the phrase as a whole is reanalyzed as an NP via the \func{Compounding} function,\footnote{Not to be confused with \emph{syntactic} nominal compounds, which feature an NP with an internal \func{Mod}.}
as in \cref{ex:LetItBe}.
If the expression is made of disjoint recognizable syntactic constituents, each attaches as \func{Compounding}, as in \cref{ex:WintersNight}.\footnote{In the future, this strategy could be extended to syntactically anomalous idioms that are not NPs, like \mention{Long time no see}. We have not encountered such cases in the corpus.}

\ex. \label{ex:LetItBe} 
    \begin{forest}
    where n children=0{
        font=\itshape, 			
        tier=word          			
      }{
      },
[NP 
    [\Det{DP}[the, roof]]
  [\Head{Nom}
    [\Head{Nom} 
       [\Head{N}[song]]]
     [\Mod{NP}
          [\Head{Nom}
            [\Node{Compounding}{VP}
                [\Head{V}[Let]]
                [\Node{Obj}{N\textsubscript{pro}}[It]]
                [\Node{Comp}{Clause}[Be, roof]]
            ]
          ]
        ]]]
    \end{forest}


\ex. \label{ex:WintersNight} 
    \begin{forest}
    where n children=0{
        font=\itshape, 			
        tier=word          			
      }{
      },
[VP 
    [\Head{V}[reading]]
    [\Obj{NP} 
      [\Head{Nom}
       [\Node{Compounding}{P}[If]]
       [\Node{Compounding}{PP}[on a Winter's Night, roof]]
       [\Node{Compounding}{NP}[a Traveler, roof]]
      ]
     ]
   ]
    \end{forest}

\subsection{Complex Numerals} \label{sec:complex-num}
Following \citet{reynolds2025} (and deviating from CGEL), we analyze compound numerals up to 99 as lexemes (\cref{sec:complex}), and complex numerals as hierarchical structures as described below.\footnote{This replaces a previous guideline to treat complex numerals as flat structures (\cref{sec:flat}).}

\subsubsection{Representation of (DP) Numerals}
Complex numerals follow these principles:
\begin{itemize}
    \item Basic lexemes are numbers 0--99 plus magnitude words (\mention{hundred}, \mention{thousand}, \mention{million}, etc.).
    \item All lexemes are always labeled as D, apart from \mention{and} and the rightmost lexeme in some cases (see \S\ref{sec:other-num}).
    \item The rightmost lexeme will be D in most cases, including where the complex numeral functions as a determiner, fused determiner--head, or fused modifier--head.
    \item Magnitude terms (hundred, million, etc.) function as heads.
    \item Factors\footnote{CGEL uses \textsc{modifier}, but that term is also used for a type of pre-determiner modifier.} modify these heads using the function \func{Mod} (Modifier factor).
    \item Additional components are joined through coordination.
    \item Coordination is usually asyndetic. \mention{And} is only possible with a basic numerative (1--99) as the rightmost constituent. We call such coordinates \textsc{additions}.
\end{itemize}

Figure~\ref{fig:complex-numeral} shows the analysis for a complex number \mention{one hundred twenty million, two hundred and ninety-nine} (100,000,299).

\begin{figure}
    \centering
    \small
    \begin{forest}
    where n children=0{
        font=\itshape, 			
        tier=word          			
      }{
      },
    [Coordination
        [\Node{Coordinate}{DP}
            [\Node{Mod}{Coordination}
                [\Node{Coordinate}{DP}
                    [\Mod{DP}[one, roof]]
                    [\Head{D},edge={line width=1pt}[hundred]]
                ]
                [\Node{Coordinate\textsubscript{add}}{DP}[twenty,roof]]
            ]
            [\Head{D},edge={line width=1pt}[million]]
        ]
        [\Node{Coordinate}{DP}
            [\Node{Mod}{DP}
                [\Head{D},edge={line width=1pt}[two]]
            ]
            [\Head{D},edge={line width=1pt}[hundred]]
        ]
        [\Node{Coordinate\textsubscript{add}}{DP}
            [\Node{Marker}{Coordinator}[and]]
            [\Head{DP},edge={line width=1pt}
                [\Head{D},edge={line width=1pt}[ninety-nine]]
            ]
        ]
    ]
    \end{forest}
    \caption{Tree diagram for \mention{one hundred twenty million, two hundred and ninety-nine} (120,000,299).} 
    \label{fig:complex-numeral}
\end{figure}

\subsubsection{Numerals in Other Categories}\label{sec:other-num}
For complex numerals with a noun or an adjective as the rightmost item, the same structural principles apply with appropriate category adjustments:

\begin{itemize}
    \item In ordinals (e.g., \mention{two million and seventh}), the final lexeme is an Adj while earlier lexemes remain D.
    \item Similarly, even when the final lexeme is a noun (e.g., \mention{millions of people}, \mention{two million is an even number}), any other lexemes are Ds.
\end{itemize}

\ex. \label{ex:ordinal-numeral}
    \begin{forest}
    where n children=0{
        font=\itshape, 			
        tier=word          			
      }{
      },
    [Coordination
        [\Node{Coordinate}{DP}
            [\Node{Mod}{DP}
                [\Head{D},edge={line width=1pt}[two]]
            ]
            [\Head{D},edge={line width=1pt}[million]]
        ]
        [\Node{Coordinate\textsubscript{add}}{AdjP}
            [\Node{Mk}{Crd}[and]]
            [\Head{Adj},edge={line width=1pt}[seventh]]
        ]
    ]
    \end{forest}

\section{Functions}
CGELBank uses the subset of the functions used in CGEL showing in \cref{fig:fxns}. Notably, CGELBank uses \func{Head} for CGEL's \func{Predicate} and \func{Predicator}.

\newcommand{\bfunc}[1]{\textbf{\func{#1}}}

\begin{figure}
\centering\small
\begin{forest}
	[{}, grow=east
		[Fused, grow=east
			[Determiner-Head (\bfunc{Det-Head})]
			[Modifier-Head (\bfunc{Mod-Head})]
			[\bfunc{Marker-Head}]
			[\bfunc{Head-Prenucleus}]
		]
		[\bfunc{Head}, grow=east]
		[Dependent, grow=east
			[Complement\\ (\bfunc{Comp}), grow=east
				[Internal, grow=east
					[Predicative Complement (\bfunc{PredComp})]
					[\bfunc{DisplacedSubj}]
					[Extraposed, grow=east
						[\bfunc{ExtraposedSubj}]
						[\bfunc{ExtraposedObj}]
					]
					[Object\\ (\bfunc{Obj}), grow=east
						[Direct (\bfunc{Obj\textsubscript{dir}})]
						[Indirect (\bfunc{Obj\textsubscript{ind}})]
					]
					[\bfunc{Particle}]
				]
				[External, grow=east
					[Extranuclear, grow=east
						[\bfunc{Prenucleus}]
						[\bfunc{Postnucleus}]
					]
					[Subject (\bfunc{Subj})]
				]
				[Indirect Complement (\bfunc{Comp\textsubscript{ind}})]
			]
			[Determiner (\bfunc{Det})]
			[Nonce (e.g.{,} \bfunc{Obj+Mod})]
			[\bfunc{Marker}]
			[\bfunc{Flat}]
			[\bfunc{Compounding}]
			[\bfunc{Coordinate}]
			[Adjunct, grow=east
				[Modifier (\bfunc{Mod}), grow=east
				]
				[\bfunc{Supplement}, grow=east
				    [\bfunc{Vocative}, grow=east]
				]
			]
		]
	]
\end{forest}
    \caption{Taxonomy of functions. The labels that appear in CGELBank are in \bfunc{bold-sans-serif}.}
    \label{fig:fxns}
\end{figure}

\subsection{Modifier vs.~Supplement}

CGEL's description of adjuncts is not entirely clear about the division between \func{modifier}s and \func{supplement}s in clause structure (see, for instance, similarities highlighted on p.~1360).
As a default, we take adjuncts of time, place, manner, condition, reason, and so on to be \func{modifier}s---even when offset by a comma. The function of \func{supplement} should be reserved for those adjuncts which are quite clearly presented as addenda: speaker commentary, clarification, parentheticals, background descriptions, and the like.

\ex. \func{Modifier}
    \a. \mention{\textbf{Because it was raining}, I was reluctant to go outside.}
    \b. \mention{\textbf{If you're hungry}, have a sandwich.}
    \b. \mention{\textbf{When the president stands}, nobody sits.}

See \cref{sec:Modifications} regarding the branching structure of modifiers. 
Note that post-head modifiers in a clause generally attach to the lowest VP.

\ex. \func{Supplement}
    \a. \mention{\textbf{As has become clear}, Sharon cannot be trusted.}
    \b. \mention{You are recommended to get a flu shot, \textbf{especially if you are over 50}.}
    \b. \mention{\textbf{Given the size of the task}, I think we can expect it to take a long time.}
    \b. \mention{\textbf{Well}, I think I'll just stay here.}
    \b. \mention{\textbf{Legally}, he's too young.}
    \b. \mention{\textbf{Remarkably}, we were not late.}
    \b. \mention{\textbf{Frankly}, I wouldn't.}
    \b. \mention{There is, \textbf{however}, a catch.}
    \b. \mention{\textbf{Damn}, what was I thinking?}
    \b. \mention{What are you up to, \textbf{Kate}?} (\func{Vocative}, a subtype of supplement)
    \b. \mention{I did it, \textbf{which was a good move}.} (relative clause as supplement)
    \b. \mention{I did it, \textbf{clearly a good move}.}

\subsection{Modifier vs.~Complement}
CGEL (starting on p.~439) regards some pre-head dependents in nominals as complements rather than attributive modifiers, as in \mention{a \underline{flower} seller} or \mention{an \underline{income tax} adviser}. In contrast, we analyze these pre-head dependents as modifiers \citep[consistent with][p.~128]{sieg2}.

\subsection{Modifiers in numerals}
The numerals before a magnitude word (e.g., \mention{two} in \mention{two hundred}) are modifiers, with the magnitude words as the head.

\subsection{Nonce functions}
Where a coordination has nonce constituents, we assign a composite function consisting of the function of each constituent within the coordinate concatenated with +. For example, in  \mention{I gave \underline{\$10 to Kim} and \underline{\$5 to Pat}}, \mention{\$10} and \mention{\$5} are objects, and \mention{to Kim} and \mention{to Pat} are complements, so the function of the coordination is \func{Obj+Comp}. In some cases, where the functions are not be consistent across coordinates, we give them with a slash notation, as in \func{Obj+PredComp/Comp}, indicating that the first coordinate includes \func{Obj} and \func{PredComp} and the second \func{Obj} and \func{Comp}.

\chapter{Tree structure and style}\label{ch:branching}
\section{Branching and labeling basics}
We use the term ``tree'' here loosely to mean a syntax tree. Strictly, the trees are directed acyclic graphs (DAGs; see \cref{sec:FoF}), and, for the most part they are true trees as defined in graph theory. The basic rules of branching in CGEL are 1) that phrases, along with non-headed constituents (see \cref{sec:non-head}), can be represented by phrase-structure trees, 2) that trees are generally right branching, 3) that binary branching is preferred, and 4) that \textit{n}-ary branching is possible. Phrases-structure trees are constructed by breaking a phrase into its constituents, which are, in turn, represented as phrase-structure trees.

Most nodes are labeled with both a category and a function. These functions being relational, they apply to the incoming branch (edge), as in (\ref{ex:edgelabels}) but are displayed at the node for convenience, as in (\ref{ex:nodelabels}). The top node on the tree has a category label only. The terminal nodes are words or, in a few cases, an en-dash `--' representing a gap.

\begin{multicols}{2}
    \ex. \label{ex:edgelabels} 
        \begin{forest}
        where n children=0{
            font=\itshape, 			
            tier=word          			
          }{
          },
        [VP
            [V,edge label={node[midway,left,font=\scriptsize]{Head}}[eat]]
            [NP,edge label={node[midway,right,font=\scriptsize]{Object}}[apples, roof]]
        ]
        \end{forest}

    \ex. \label{ex:nodelabels} 
        \begin{forest}
        where n children=0{
            font=\itshape, 			
            tier=word          			
          }{
          },
            [VP
                [\Head{V}[eat]]
                [\Obj{NP}[apples, roof]]
            ]
        \end{forest}
        
\end{multicols}

\subsection{Unary branching}
A clause like \mention{stop} is headed by a VP, which is headed by a V. This is an example of unary branching. CGEL often omits phrasal nodes in unary branches. For example, the \func{determiner} in \mention{some children} is represented as a D instead of a DP in CGEL's [13] (p.~26), and no Nom is shown between the NP and the N. This tree from CGEL is reproduced in (\ref{ex:Det-D}). 

\ex. \label{ex:Det-D} 
    \begin{forest}
    where n children=0{
        font=\itshape, 			
        tier=word          			
      }{
      },
        [NP
            [\Det{D}[some]]
            [\Head{N}[children]]
        ]
    \end{forest}

In CGELBank, nodes are never omitted, so that we represent the same NP with unary branches, as in (\ref{ex:Det-DP}).\footnote{In this guide, we sometimes omit internal structure, but this is always indicated by a triangle in the tree.}

\ex. \label{ex:Det-DP}
    \begin{forest}
    where n children=0{
        font=\itshape, 			
        tier=word          			
      }{
      },
        [NP
            [\Det{DP}
                [\Head{D}[some]]
            ]
            [\Head{Nom}
                [\Head{N}[children]]
            ]
        ]
    \end{forest}


A single modifier may be the sibling of a lexical \func{head}. This is exemplified in \cref{ex:compoundNNN}, where each modifier (which may
itself be a Nom constituent) attaches at a Nom. 
The modifier nearest to the head shares the Nom projected by the head, 
while remaining modifiers add extra Nom layers.
Note that per the Lexical Projection Principle (\cref{sec:lexprojection}), each modifier lexeme projects its own Nom level, which is never shared with the head.

\ex. \label{ex:compoundNNN}
    \begin{forest}
    where n children=0{
        font=\itshape, 			
        tier=word          			
      }{
      },
      [NP
      [\Head{Nom}
        [\Mod{Nom}
            [\Head{N}[desert]]]
        [\Head{Nom}
            [\Mod{Nom}
                [\Head{N}[weather]]
            ]
            [\Head{N}[stations]]
        ]]]
    \end{forest}

In \cref{ex:Det-DP,ex:compoundNNN}, for every word except \mention{stations}, the phrasal constituent projected by the lexical node is unary. Likewise, an intransitive unmodified V may be the sole element of a VP. Coordinations of intransitive unmodified verbs must bear the unary VP layer as well (even in the head of a marked coordinate; note that coordination in CGELBank is always between phrasal categories: \cref{sec:lexcoord}).

Every clause must be headed by a VP (or another clause level), and every NP must be headed by a Nom (or another NP level). Clause and NP constituents may be unary to conform to this requirement.\footnote{The term ``clause'' here covers constituents of both the Clause and \Clauserel categories.}

\subsection{Binary branching}\label{sec:Modifications}

\textbf{Most headed nonlexical constituents exhibit binary branching.} If a phrase of type XP contains multiple dependents, these will generally be layered so as to attach one at a time going outward from the head, forming intermediate XP constituents. 
A typical example is \cref{ex:XP}, where the VP-internal complement (\func{Obj}) forms a VP constituent with the head verb, and that VP heads a larger VP constituent with a modifier.

\ex. \label{ex:XP} 
    \begin{forest}
    where n children=0{
        font=\itshape, 			
        tier=word          			
      }{
      },
        [VP
            [\Mod{AdvP}[quickly, roof]]
            [\Head{VP}
                [\Head{V}[eat]]
                [\Obj{NP}[apples, roof]]
            ]
        ]
    \end{forest}
    
Exceptions to binary branching are discussed in \cref{sec:beyondbinary}.

\subsubsection{Modification} 

Modification is always binary: a constituent attaching as \func{Mod} always has \func{Head} as its sibling.\footnote{Binary not counting supplements: \cref{sec:Supplements}.} 
A \func{modifier} will never be sibling to a complement, for example.

Where there are multiple \func{modifiers}, these are layered, one per XP. Generally those farther from the \func{head} are higher in the tree structure, but semantics are also brought to bear in adjudicating when there are both pre- and post-head \func{modifiers}.

Post-head \func{modifiers} attach as low as possible in the tree structure. For example, in principle, \mention{quickly} in \mention{It can run quickly} could attach in the \mention{can} clause, the \mention{can} VP, the \mention{run} clause, or the \mention{run} VP. Unless there is a clear reason to do otherwise, we attach it to the \mention{run} VP.


One upshot of this preference for binary branching is that, while it is true in principle that, for example, a VP headed by a transitive verb licenses an \func{object} and permits certain kinds of modifiers, it is not the case that a \func{modifier} or \func{complement} may always be branched from any VP. This also applies to an XP \func{coordinate} with a \func{marker} (see \cref{sec:Supplements}).


\subsubsection{Post-head pre-complement modifiers}
Should a \func{modifier} come between a \func{head} and its \func{complement}, there are two possible analyses. The first is that the \func{complement} is post-posed (see \cref{sec:Postposing}), usually because it is heavy. In such a case, we include a gap in the immediate post-head position in an XP with the \func{head}. The \func{modifier} is attached to a higher XP, and the post-posed \func{complement} to a yet higher XP.

In some cases, though, a \func{modifier} naturally comes between a \func{head} an its \func{complement}, so the inner VP will contain the \func{modifier} whereas the outer VP will add the \func{complement}. We see this most commonly with auxiliary verbs (e.g., \mention{You may not go}). In such cases, we take \mention{not} to form a VP with \mention{may}, and this VP takes the clause \mention{go} as its \func{complement}.
A non-auxiliary example attested in CGELBank is \mention{think again about\dots}, where the \mention{about}-PP is licensed by the verb.

\subsection{Beyond binary branching}\label{sec:beyondbinary}\label{sec:Supplements}

\subsubsection{Morphologically derived expressions}
Complex expressions such as personal names may contain two or more \func{Flat} or \func{Compounding} dependents: see \cref{sec:derived}.

\subsubsection{Coordinations}
Coordination constituents are non-headed and may have more than two children in \func{Coordinate} function. See \cref{ch:coord}.

\subsubsection{VP-internal complements}
While multiple \func{modifiers} are layered, multiple (internal, non-extraposed) \func{complements} within a VP will typically attach on the same level. A typical example of ternary branching would be in a ditransitive construction like (\ref{ex:ditran}).

\ex. \label{ex:ditran} 
    \begin{forest}
    where n children=0{
        font=\itshape, 			
        tier=word          			
      }{
      },
        [VP
            [\Head{V}[give]]
            [\Node{Obj\textsubscript{ind}}{NP}[us, roof]]
            [\Node{Obj\textsubscript{dir}}{NP}[the apples, roof]]
        ]
    \end{forest}

Extraposed complements trigger a separate level: see \cref{sec:extraposition}.

The treatment of complements within VPs is exceptional. In NP structure, complements and modifiers are treated the same: each attaches within a separate Nom layer.

\subsubsection{PP-internal complements}

We follow CGEL (p.~641) in recognizing double-complement PPs such as the following:

\ex. \label{ex:dblcomppp} 
    \begin{forest}
    where n children=0{
        font=\itshape, 			
        tier=word          			
      }{
      },
        [PP
            [\Head{P}[from]]
            [\Node{Obj}{NP}[Boston, roof]]
            [\Node{Comp}{PP}[to Providence, roof]]
        ]
    \end{forest}

\subsubsection{Supplements} 
CGEL analyzes \func{supplements} as not being fully integrated into phrase structure. To illustrate this, it shows them as separate trees with an arrow pointing to the supplement's semantic anchor as in \cref{ex:CGELSupp} (CGEL's [12], p.~1354).

\ex. \label{ex:CGELSupp}
    \includegraphics[scale=0.5]{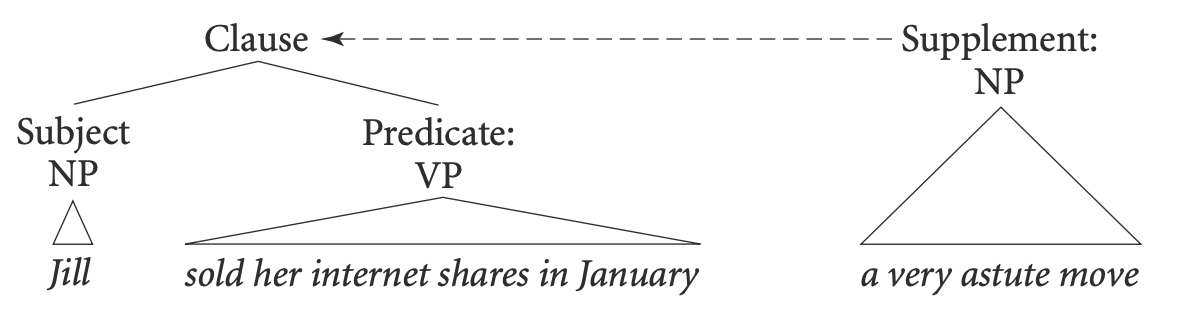}

Without disagreeing with the analysis in CGEL, CGELBank presents supplements as an additional branch from the anchor for simplicity, as in \cref{ex:SuppBasic}.

\ex. \label{ex:SuppBasic} 
\begin{forest}
where n children=0{
    font=\itshape, 			
    tier=word          			
  }{
  },
    [Clause
        [\Subj{NP}[Jill, roof]]
    	[\Head{VP}[sold her internet shares in January, roof]]
    	[\Sup{NP}[a very astute move, roof]]
    	[\Sup{AdvP}[frankly, roof]]
    ]
\end{forest}

\subsection{Summary of constraints on branching}

Putting together the branching principles expressed above and in the Lexical Projection Principle (\cref{sec:lexprojection}), and setting aside non-headed constituents, we have the following constraints:

\begin{enumerate}
\item A node of a lexical category that projects a constituent of a corresponding phrasal category must do so (as its head) unless it is \Vaux in \func{Prenucleus} function.

\item A lexical node has too many layers if its parent is in \func{Head} function, the categories of its parent and grandparent are the same, and the grandparent is not functioning as a \func{Coordinate}.

\item In a VP, constituents in non-extraposed internal complement\footnote{Non-extraposed internal complements are constituents in the function of \func{Comp}, \func{Obj}, \func{Obj\textsubscript{dir}}, \func{Obj\textsubscript{ind}}, \func{PredComp}, \func{Particle}, or \func{DisplacedSubj}. CGEL makes a further distinction of \emph{core} vs.~\emph{non-core} internal complements, but this distinction is not currently reflected in CGELBank.} functions will be on the same level except where separated from the head by an intervening modifier. Every modifier and extraposed complement forms a binary VP with the head (not counting supplements).

In other words, each non-unary VP level consists of a \func{Head} plus a)~a single \func{Mod}, b)~an extraposed complement, or c)~any number of non-extraposed internal complements (and there should not be two consecutive levels of type (c)).

\item An NP must be headed by a Nom or another NP level.

\item A Clause or \Clauserel must be headed by a VP or another Clause or \Clauserel level. 

\item A phrasal (headed) constituent other than VP must be no more than binary (not counting supplements).

\item A unary phrase (not counting supplements) should not be headed by a constituent with the same phrasal category as this would be a vacuous branch---though an exception is necessary for the intermediate node of fused structures described below.

\end{enumerate}

\section{Fusion of functions} \label{sec:FoF}
CGEL departs from strict tree structure in cases where a constituent functions both as a \func{head} and as a dependent while still being a directed acyclic graph (DAG), as detailed in \citep{pullum-09}.
A constituent has two parent constituents with respect to which it bears different functions (one of them \func{head}); these functions are shown hyphenated in a single label.
Typically, the constituent with two parents is deeper in the tree by one level along one branch than along the other, though the difference may involve multiple levels as in \cref{ex:fused-2postmods} below.

\subsection{Determiner-Head}

\begin{multicols}{2}
\ex. \label{ex:det-head} 
\begin{forest}
where n children=0{
    font=\itshape, 			
    tier=word          			
  }{
  },
    [NP,s sep=-1.5em
		[\phantom{X}\hspace*{-2em},tier=dh]
		[\small\textsf{Det-Head:}\\DP,no edge,tier=dh
			[this, roof]
		]
		[\small\textsf{Head:}\\Nom
			[\hspace*{-2em}\phantom{X},tier=dh]
		]
	]
\end{forest}

\ex. \label{ex:fusedAnyone} 
\begin{forest}
where n children=0{
    font=\itshape, 			
    tier=word          			
  }{
  },
    [NP,s sep=-1.3em
    	[\phantom{X}\hspace*{-2em},tier=dh2]
    	[\small\textsf{Det-Head:}\\DP,no edge,tier=dh2
    		[hardly anyone, roof]
    	]
    	[\small\textsf{Head:}\\Nom
    		[\hspace*{1em}\phantom{X},tier=dh2]
    		[\Mod{AdjP}[present, roof]]
    	]
    ]
\end{forest}

\ex.\label{ex:fused-2postmods}
\begin{forest}
where n children=0{
    font=\itshape, 			
    tier=word          			
  }{
  },
    [NP
        [\Head{Nom},before drawing tree={x+=1.5em}
            [\Head{Nom},before drawing tree={x+=2.5em}
            	[\Node{Det-Head}{DP},no edge[everyone, roof]] { \draw[-] (!uuu.south) -- (); \draw[-] (!u.south) -- (); }
        		[\Mod{AdjP}[present, roof]]
            ]
            [\Mod{\Clauserel}[that I knew, roof]]
    	]
    ]
\end{forest}

\ex. \label{ex:partitive} 
\begin{forest}
where n children=0{
    font=\itshape, 			
    tier=word          			
  }{
  },
    [NP,s sep=-2em
		[\phantom{X}\hspace*{-2em},tier=dh2]
		[\small\textsf{Det-Head:}\\DP,no edge,tier=dh2
			[some, roof]
		]
		[\small\textsf{Head:}\\Nom
			[\hspace*{1em}\phantom{X},tier=dh2]
			[\Comp{PP}
				[\Head{P}[of]]
				[\Obj{NP}[those, roof]]
			]
		]
	]
\end{forest}

\end{multicols}

\subsection{Modifier-Head}

\ex. \label{ex:mod-head} 
\begin{forest}
where n children=0{
    font=\itshape, 			
    tier=word          			
  }{
  },
    [Clause
    	[\Subj{NP}
    	    [\Det{DP}[the, roof]]
    	    [\Head{Nom},s sep=-1.5em
    			[\phantom{X}\hspace*{-4em},tier=dh]
    			[\small\textsf{Mod-Head:}\\AdjP,no edge,tier=dh
    				[rich, roof]
    			]
    			[\small\textsf{Head:}\\Nom
    				[\hspace*{-4em}\phantom{X},tier=dh]
    			]
    		]
    	]
    	[\Head{VP}[get richer, roof]]
    ]
\end{forest}

\subsection{Head-Prenucleus}\label{sec:head-prenuc}

The \func{Head-Prenucleus} function appears in fused relative constructions. 
(For non-fused relatives, see \cref{sec:relatives}.)

\subsubsection{Relative NP}

\ex. \label{ex:relNP} 
\begin{forest}
where n children=0{
    font=\itshape, 			
    tier=word          			
  }{
  },
    [NP
      	[\Head{Nom},s sep=-1em
      		[\phantom{X}\hspace*{-2em},tier=dh2]
      		[\small\textsf{Head-Prenucleus:}\\NP\textsubscript{x},no edge,tier=dh2,s sep=-2em
      				[whichever ones, roof]
      		]
      		[\small\textsf{Mod:}\\ \Clauserel
      			[\hspace*{1em}\phantom{X},tier=dh2]
      			[\Head{\Clauserel}
      				[\Subj{NP}[you, roof]]
      				[\Head{VP}
      					[\Head{V}[want]]
      					[\Obj{GAP\textsubscript{x}}[--]]
      				]
      			]
      		]
      	]
      ]
\end{forest}

\subsubsection{Relative PP}

Though relative clauses do not typically function as modifiers in PPs, this is the analysis in CGEL and also in \citet{payne-07}.

\ex. \label{ex:relPP} 
\begin{forest}
where n children=0{
    font=\itshape, 			
    tier=word          			
  }{
  },
    [PP,s sep=-1em
      	[\phantom{X}\hspace*{-2em},tier=dh2]
      	[\small\textsf{Head-Prenucleus:}\\PP\textsubscript{x},no edge,tier=dh2,s sep=-2em
      	    [wherever, roof]
      	]
      	[\small\textsf{Mod:}\\ \Clauserel
      		[\hspace*{1em}\phantom{X},tier=dh2]
      		[\Head{\Clauserel}
      			[\Subj{NP}[you, roof]]
      			[\Head{VP}
      				[\Head{V\textsubscript{aux}}[are]]
      				[\Comp{GAP\textsubscript{x}}[--]]
      			]
      		]
      	]
    ]
\end{forest}

\subsubsection{Relative AdjP}

\ex. \label{ex:relAdjP} 
\begin{forest}
where n children=0{
    font=\itshape, 			
    tier=word          			
  }{
  },
    [AdjP,s sep=-1em
      	[\phantom{X}\hspace*{-2em},tier=dh2]
      	[\small\textsf{Head-Prenucleus:}\\AdjP\textsubscript{x},no edge,tier=dh2,s sep=-2em
      	    [however small, roof]
      	]
      	[\small\textsf{Mod:}\\ \Clauserel
      		[\hspace*{1em}\phantom{X},tier=dh2]
      		[\Head{\Clauserel}
      			[\Subj{NP}[it, roof]]
      			[\Head{VP}
      				[\Head{V\textsubscript{aux}}[is]]
      				[\PredComp{GAP\textsubscript{x}}[--]]
      			]
      		]
      	]
    ]
\end{forest}

\subsubsection{Relative AdvP}

\ex. \label{ex:relAdvP} 
\begin{forest}
where n children=0{
    font=\itshape, 			
    tier=word          			
  }{
  },
    [AdvP,s sep=-1em
      	[\phantom{X}\hspace*{-2em},tier=dh2]
      	[\small\textsf{Head-Prenucleus:}\\AdvP\textsubscript{x},no edge,tier=dh2,s sep=-2em
      	    [however often, roof]
      	]
      	[\small\textsf{Mod:}\\ \Clauserel
      		[\hspace*{1em}\phantom{X},tier=dh2]
      		[\Head{\Clauserel}
      			[\Subj{NP}[I, roof]]
      			[\Head{VP}
      				[\Head{V}[try]]
      				[\Mod{GAP\textsubscript{x}}[--]]
      			]
      		]
      	]
    ]
\end{forest}

\subsection{Marker-Head}\label{sec:marker-head}
We have two cases of a phrase in \func{Marker-Head} function: The case of \mention{etc.}\ (see \cref{sec:Etc}) and the case of elliptical stranding of \mention{to} (see \cref{sec:ellipstrand}).

\section{Indirect complements}\label{sec:compind}
An indirect complement forms a binary branch with the lowest possible XP. In (\ref{ex:Indirect}), the \func{Comp\textsubscript{ind}} is licensed by \mention{so}.

\ex. \label{ex:Indirect} 
\begin{forest}
where n children=0{
    font=\itshape, 			
    tier=word          			
  }{
  },
[NP
	[\Head{NP}
		[\Mod{AdjP}[so great, roof]]
		[\Head{NP}[a loss]]
	]
	[\Node{Comp\textsubscript{ind}}{Clause}[that we gave up, roof]]
]
\end{forest}

See \cref{sec:hollow} for examples of indirect complements with gaps.

\section{Predicative complements}\label{sec:predcomp}

A constituent in predicative complement function (licensed by a verb like \mention{be}, \mention{seem}, or the second internal complement of \mention{consider}) is typically an NP or AdjP.\footnote{The sentence \mention{I consider him to be handsome.}\ is a paraphrase of \cref{ex:predcomp-vp} in which \mention{to be handsome} is a catenative---not predicative---complement of \mention{consider}, thus simply \func{Comp}; and the subordinate clause contains an AdjP in \func{PredComp} function.}
The \func{PredComp} function may occur directly under a VP, as in \cref{ex:predcomp-vp}. It may also be marked by the preposition \mention{as}. We adopt the structure in \cref{ex:predcomp-pp} following \citet[p.~194]{sieg2}.

\begin{multicols}{2}
\ex. \label{ex:predcomp-vp} 
\begin{forest}
where n children=0{
    font=\itshape, 			
    tier=word          			
  }{
  },
[Clause
	[\Subj{NP}[I, roof]]
	[\Head{VP}
        [\Head{V}[consider]]
		[\Obj{NP}[him, roof]]
        [\PredComp{AdjP}[handsome, roof]]
	]
]
\end{forest}

\ex. \label{ex:predcomp-pp} 
\begin{forest}
where n children=0{
    font=\itshape, 			
    tier=word          			
  }{
  },
[Clause
	[\Subj{NP}[I, roof]]
	[\Head{VP}
        [\Head{V}[regard]]
		[\Obj{NP}[him, roof]]
        [\Comp{PP}
            [\Head{P}[as]]
            [\PredComp{AdjP}[handsome, roof]]]
	]
]
\end{forest}

\end{multicols}

As an example of an NP functioning as \func{PredComp}, replace \mention{handsome} in \cref{ex:predcomp-vp,ex:predcomp-pp} with \mention{a~great teacher}.

\section{Information packaging clause constructions}
\subsection{Pre- and Postposing}

CGEL claims that ``subject postposing affects order, not function'' (p.~244), but this is not strictly true, at least not in the CGELBank tree structure. The pre- or postposed constituent is co-indexed to a gap with the function the constituent would have had in the basic position, but the constituent itself is a pre- or postnucleus.

\ex. \label{ex:Posposing} 
\begin{forest}
where n children=0{
    font=\itshape, 			
    tier=word          			
  }{
  },
[Clause
	[\Head{Clause}
		[\Subj{NP}[he, roof]]
		[\Head{VP}
			[\Head{V}[gave]]
			[\Obj{GAP\textsubscript{x}}[--]]
			[\Comp{PP}[to charity, roof]]
		]
	]
	[\Node{Postnucleus}{NP\textsubscript{x}}[everything he had earned from the years of toil, roof]]
]
\end{forest}

Additional examples appear in \cref{sec:pre}.

\subsection{Inversion}\label{sec:inversion}

The auxiliary verb is a \func{Prenucleus} co-indexed to a gap in the usual location. If the inversion is triggered by a fronted element, it is also a \func{Prenucleus}.

\ex. \label{ex:Inversion} 
\begin{forest}
where n children=0{
    font=\itshape, 			
    tier=word          			
  }{
  },
	[Clause
		[\Node{Prenucleus}{AdvP\fbox{\textsubscript{x}}}[thus, roof]]
		[\Head{Clause}
			[\Node{Prenucleus}{V\textsubscript{aux}\fbox{\textsubscript{y}}}[had]]
			[\Head{Clause}
				[\Subj{NP}[they, roof]]
				[\Head{VP}
					[\Head{GAP\fbox{\textsubscript{y}}}[--]]
					[\Comp{Clause}
                        [\Head{VP}
    						[\Head{V}[parted]]
    						[\Mod{GAP\fbox{\textsubscript{x}}}[--]]
    					]
                    ]
				]
			]
		]
	]
\end{forest}

Additional examples and discussion appear in \cref{sec:AdjInv}.

\subsection{Existential clauses}

The \func{displaced subject} is a complement in the VP along with any others such as a \func{PredComp}.

\ex. \label{ex:Existential} 
\begin{forest}
where n children=0{
    font=\itshape, 			
    tier=word          			
  }{
  },
	[Clause
		[\Subj{NP}[there, roof]]
		[\Head{VP}[\Head{V\textsubscript{aux}}[is]]
			[\Node{DisplacedSubj}{NP}[something, roof]]
			[\PredComp{AdjP}[\Head{Adj}[wrong]]]
		]
	]
\end{forest}

\subsection{Extraposition}\label{sec:extraposition}
In CGELBank, \func{extraposed subject} is located in an outer VP layer \citep[in line with][p.~372]{sieg2}:

\ex. \label{ex:Extraposed} 
\begin{forest}
where n children=0{
    font=\itshape, 			
    tier=word          			
  }{
  },
    [Clause
        [\Subj{NP}[it, roof]]
        [\Head{VP}
        	[\Head{VP}
        		[\Head{V\textsubscript{aux}}[is]]
        		[\PredComp{AdjP}[hard, roof]]
        	]
    	[\Node{ExtraposedSubj}{Clause}[to keep it up, roof]]
    	]
    ]
\end{forest}

\subsection{Cleft clauses}
The relative clause appears in extranuclear position at the end (p.~1416). Note that this \func{postnucleus} is not coindexed.

\ex. \label{ex:Cleft} 
\begin{forest}
where n children=0{
    font=\itshape, 			
    tier=word          			
  }{
  },
[Clause
	[\Head{Clause}
		[\Subj{NP}[it, roof]]
		[\Head{VP}
			[\Head{V\textsubscript{aux}}[was]]
			[\PredComp{NP\textsubscript{x}}[a bee, roof]]
		]
	]
	[\Node{Postnucleus}{Clause\textsubscript{rel}}
		[\Mk{Sdr}[that]]
		[\Head{Clause}
			[\Subj{GAP\textsubscript{x}}[--]]
			[\Head{VP}[stung me, roof]]
		]
	]
]
\end{forest}

Occasionally, a nonfinite clause will appear instead of a relative clause (p.~1420). Without a relative clause, there is no gap in the \func{postnucleus}. Here is one in an interrogative sentence:

\ex. \label{ex:NonfiniteCleftQuestion} 
\begin{forest}
where n children=0{
    font=\itshape, 			
    tier=word          			
  }{
  },
[Clause
    [\Node{Prenucleus}{V\textsubscript{aux}\fbox{\textsubscript{x}}}[was]]
	[\Head{Clause}
        [\Head{Clause}
    		[\Subj{NP}[it, roof]]
    		[\Head{VP}
    			[\Head{GAP\fbox{\textsubscript{x}}}[--]]
    			[\PredComp{NP}[a bee, roof]]
    		]
    	]
    	[\Node{Postnucleus}{Clause}
    		[\Head{VP}[making all that noise, roof]
    		]
    	]
    ]
]
\end{forest}

\subsection{Passives}\label{sec:pass}\label{sec:preppass}
Any internalised complement is a \func{Comp} in the VP.
\ex. \label{ex:Passive} 
    \begin{forest}
    where n children=0{
        font=\itshape, 			
        tier=word          			
      }{
      },
    [Clause
    	[\Subj{NP}[it, roof]]
    	[\Head{VP}
    		[\Head{V\textsubscript{aux}}[was]]
    		[\Comp{Clause}
    			[\Head{VP}
    				[\Head{V}[followed]]
    				[\Comp{PP}[by a comma, roof]]
    			]
    		]
    	]
    ]
    \end{forest}

Prepositional passives are distinguished from regular passives by stranding (\cref{sec:prepstrand}):
\ex. \label{ex:PrepPassive} 
    \begin{forest}
    where n children=0{
        font=\itshape, 			
        tier=word          			
      }{
      },
    [Clause
    	[\Subj{NP}[it, roof]]
    	[\Head{VP}
    		[\Head{V\textsubscript{aux}}[was]]
    		[\Comp{Clause}
    			[\Head{VP}
    				[\Head{V}[touched]]
                    [\Comp{PP\textsubscript{strand}}[
                        \Head{P}[on]
                    ]]
    				[\Comp{PP}[by the analysis, roof]]
    			]
    		]
    	]
    ]
    \end{forest}

\subsection{Dislocation}
The dislocated constituent is a \func{Supplement}.
\ex. \label{ex:Dislocation} 
    \begin{forest}
    where n children=0{
        font=\itshape, 			
        tier=word          			
      }{
      },
    [Clause
    	[\Sup{NP}[me, roof]]
    	[\Subj{NP}[I, roof]]
    	[\Head{VP}[wouldn't do it, roof]]
    ]
    \end{forest}

\chapter{Gaps and co-indexing}
    \section{Gapping basics}\label{sec:gapping-basics}
Formal syntactic theories have been known to postulate elaborate inventories of null elements and systems of movement in order to account for phenomena such as ellipsis and control.
CGEL positions itself as a descriptive framework, and as a rule, avoids invisibilia---but unbounded dependencies and other noncanonical word order constructions are the exception. 
For such constructions, a \textbf{gap} node appears as a leaf in the tree to indicate the canonical position of a constituent, and the gap is coindexed with the overt constituent in its ``surface'' position.
This makes for a fairly intuitive description of sentences with relative clauses, WH-questions, inversion, and pre-\slash postposed elements.

CGEL notates gaps with a `\_\_', but we use an en-dash `--'. Gap nodes are labeled in the typical manner with a function and with ``GAP'' appearing as the category. Gaps are co-indexed with a subscript such as x to any extranuclear node or, where no such node exists, 
to an antecedent. A typical example is shown in (\ref{ex:basicgap}).

\ex. \label{ex:basicgap} 
    \begin{forest}
    where n children=0{
        font=\itshape, 			
        tier=word          			
      }{
      },
        [NP
            [\Det{DP}[the, roof]]
            [\Head{Nom}
                [\Head{Nom}[\Head{N}[person]]]
                [\Mod{Clause\textsubscript{rel}}
                    [\Node{Prenucleus}{NP\textsubscript{x}}[who, roof]]
                    [\Head{Clause\textsubscript{rel}}
                        [\Subj{NP}[I, roof]]
                        [\Head{VP}
                            [\Head{V}[met]]
                            [\Obj{GAP\textsubscript{x}}[--]]
                        ]
                    ]
                ]
            ]
        ]
    \end{forest}

A gap may even appear in extra-nuclear material, as in GAP\textsubscript{y} in (\ref{ex:Steedman}) from Mark Steedman.

\ex. \label{ex:Steedman}
\begin{forest}
where n children=0{
    font=\itshape, 			
    tier=word          			
  }{
  },
	[Clause
		[\Node{Prenucleus}{Clause\textsubscript{x}}
			[\Node{Prenucleus}{NP\textsubscript{y}}[whose woods, roof]]
			[\Head{Clause}
				[\Subj{NP}[these, roof]]
				[\Head{VP}
					[\Head{V\textsubscript{aux}}[are]]
					[\PredComp{GAP\textsubscript{y}}[--]]
				]
			]
		]
		[\Head{Clause}
 			[\Subj{NP}[I, roof]]
 			[\Head{VP}
 				[\Head{V}[think]]
	  			[\Comp{Clause}
	  				[\Subj{NP}[I, roof]]
  				 [\Head{VP}
	  				 	[\Head{V}[know]]
	  				 	[\Comp{GAP\textsubscript{x}}[--]]
  				 ]
	  			]
	  		]
	  	]
  	]
\end{forest}

A gap can also be co-indexed to a gap.
\mention{Which\textsubscript{x} was it --\textsubscript{x} [she put it in --\textsubscript{x}]?}

\paragraph{Formal constraints.} In CGELBank, we enforce the following requirements:
\begin{itemize}
    \item Every gap must be coindexed to exactly one overt element in the sentence (and potentially to other gaps).
    \begin{itemize}
        \item The overt element should bear a nonlexical category where possible. For example, interrogative or relative \mention{who} should be coindexed at the NP level. Exceptions where coindexation is at the lexical level: the \Vaux in \func{Prenucleus} function with subject-auxiliary inversion \cref{ex:subjaux}; and an N that is sister to a bare relative clause modifier \cref{ex:barerelative} or modifier containing a hollow clause \cref{ex:HollowForPurpose}. 
    
    \end{itemize}
    \item Every distinct coindexation variable in the sentence must apply to at least one gap.
    \item No overt element or gap may receive more than one coindexation variable (e.g., a noun with two non-WH relative clause modifiers will have the two coindexation variables at different Nom levels).
\end{itemize}
Thus, constituents involved in coreference phenomena do not receive coindexation except to resolve a gap. This rules out (in most cases) ordinary anaphora, reflexive anaphors, resumptive pronouns, and nouns that are relativized with an overt relative pronoun (\cref{sec:relatives}).

In addition to these hard constraints, we point out a tendency:
\begin{itemize}
\item A gap will \emph{usually} have at least one non-gap sister in a function other than \func{supplement} (i.e.~it is not the child of a unary rule, and a binary rule does not consist simply of two gaps).
\end{itemize}
However, exceptions do exist, at least in the binary case. \Cref{ex:doublegap} shows an open interrogative copula clause, where the interrogative phrase \mention{how tall} is fronted from predicative complement position and the copula is fronted in subject-auxiliary inversion, leaving a VP with two gaps and no overt constituents:

\ex. \label{ex:doublegap} 
    \begin{forest}
where n children=0{
    font=\itshape, 			
    tier=word          			
  }{
  },
[Clause
	[\Node{Prenucleus}{NP\fbox{\textsubscript{x}}}[how tall, roof]]
	[\Head{Clause}
		[\Node{Prenucleus}{V\textsubscript{aux}\fbox{\textsubscript{y}}}[is]]
		[\Head{Clause}
			[\Subj{NP}[that tree, roof]]
			[\Head{VP}
				[\Head{GAP\fbox{\textsubscript{y}}}[--]]
				[\PredComp{GAP\fbox{\textsubscript{x}}}[--]]
			]
		]
	]
]
\end{forest}


\section{Pre- and postposing} \label{sec:pre}
A pre- or postposed dependent is indicated with a gap in the basic position co-indexed to the pre- or postposed element. 



\subsection{Preposing}
\ex. \label{ex:preposing} 
    \begin{forest}
where n children=0{
    font=\itshape, 			
    tier=word          			
  }{
  },
[Clause
	[\Node{Prenucleus}{Clause\textsubscript{x}}[the rest, roof]]
	[\Head{Clause}
		[\Subj{NP}[I, roof]]
		[\Head{VP}
			[\Head{V}[keep]]
			[\Comp{GAP\textsubscript{x}}[--]]
		]
	]
]
\end{forest}

\subsection{Postposing} \label{sec:Postposing}
\ex.\label{ex:postposing} 
Postposing due to weight (pp.~1382--1383):

    \begin{forest}
where n children=0{
    font=\itshape, 			
    tier=word          			
  }{
  },
[Clause
	[\Head{Clause}
		[\Subj{NP}[we, roof]]
		[\Head{VP}
			[\Head{VP}
				[\Head{V}[hold]]
				[\Comp{GAP\textsubscript{x}}[--]]
			]
			[\Mod{PP}[for posterity, roof]]
		]
	]
	[\Node{Postnucleus}{Clause\textsubscript{x}}[all the archaic material that has been unearthed, roof]]
]
\end{forest}

\needspace{10em}
\ex.\label{ex:posing} 
Subject postposing with verb of reporting (p.~1384):

    \begin{forest}
where n children=0{
    font=\itshape, 			
    tier=word          			
  }{
  },
[Clause
	[\Node{Prenucleus}{Clause\textsubscript{y}}[it works, roof]]
	[\Head{Clause}
		[\Head{Clause}
			[\Subj{GAP\textsubscript{x}}[--]]
			[\Head{VP}
				[\Head{V}[said]]
				[\Comp{GAP\textsubscript{y}}[--]]
			]
		]
		[\Node{Postnucleus}{NP\textsubscript{x}}[Jim, roof]]
	]
]
\end{forest}

\subsection{Adjuncts in clause structure} \label{sec:AdjInv}
In CGELBank, pre-clause \func{adjuncts} are not said to be preposed -- and thus do not trigger a gap\footnote{Contrast CGEL's \mention{\uline{If you pay me}}\textsubscript{i}, \mention{I'll do it --}\textsubscript{i} (p.~1092).} -- except where the \func{adjunct} triggers \func{subject}-auxiliary inversion as in the following cases:
\begin{itemize}
    \item A negative polarity item (e.g., \mention{\underline{Never} had I seen the like.})
    \item Phrases starting with \mention{only} (e.g., \mention{\underline{Only once} had I seen it.})
    \item Other adjuncts, such as \mention{thus} or \mention{yet} (e.g., \mention{\underline{Thus} had they parted.})
\end{itemize}
See \cref{sec:inversion} for an example tree.

\section{Subject–auxiliary inversion}\label{sec:sai}
Subject–auxiliary inversion (SAI) is a gapped construction with a co-indexed element (typically a sole \Vaux) in the prenucleus. This applies equally to \mention{do} support. A typical example is shown in \cref{ex:subjaux}.

\ex. \label{ex:subjaux} 
    \begin{forest}
    where n children=0{
        font=\itshape, 			
        tier=word          			
      }{
      },
    [Clause
    	[\Node{Prenucleus}{V\textsubscript{aux}\fbox{\textsubscript{x}}}[were]]
    	[\Head{Clause}
    		[\Subj{NP}[you, roof]]
    		[\Head{VP}
    			[\Head{GAP\fbox{\textsubscript{x}}}[--]]
    			[\PredComp{Adj}[okay, roof]]
    		]
    	]
    ]
    \end{forest}

Note that the prenucleus in SAI has no VP, just a single lexical item. In rare circumstances, a larger constituent must be created for the prenucleus:

\ex. \label{ex:subjauxlarge} 
    \begin{forest}
    where n children=0{
        font=\itshape, 			
        tier=word          			
      }{
      },
    [Clause
    	[\Node{Prenucleus}{Coordination\textsubscript{x}}[will or won't, roof]]
    	[\Head{Clause}
    		[\Subj{NP}[you, roof]]
    		[\Head{VP}
    			[\Head{GAP\textsubscript{x}}[--]]
    			[\Comp{Clause}[do it, roof]]
    		]
    	]
    ]
    \end{forest}

(See \cref{ex:lexemes6} for the structure of the coordination of auxiliaries.)

\section{Subject–dependent inversion}\label{sec:sdi}
Subject–dependent inversion (SDI) is a double-gapped construction with a subject gap co-indexed to the postnucleus and another gap (typically a complement of the head verb) co-indexed to the prenucleus (p.~1385). A typical example is shown in \cref{ex:subjdep}.

\ex. \label{ex:subjdep} 
    \begin{forest}
where n children=0{
    font=\itshape, 			
    tier=word          			
  }{
  },
[Clause
	[\Node{Prenucleus}{PP\textsubscript{y}}[here, roof]]
	[\Head{Clause}
		[\Head{Clause}
			[\Subj{GAP\textsubscript{x}}[--]]
			[\Head{VP}
				[\Head{V\textsubscript{aux}}[is]]
				[\Comp{GAP\textsubscript{y}}[--]]
			]
		]
		[\Node{Postnucleus}{NP\textsubscript{x}}[Jim, roof]]
	]
]
\end{forest}

\section{Unbounded dependencies}

\subsection{Relative constructions} \label{sec:relatives}
Every relative construction includes a clause with a gap that is co-indexed to a prenucleus or antecedent. This includes \func{subject} relatives such as (\ref{ex:subjgap}).

\ex. \label{ex:subjgap} 
    \begin{forest}
where n children=0{
    font=\itshape, 			
    tier=word          			
  }{
  },
[NP
	[\Det{DP}[a, roof]]
	[\Head{Nom}
		[\Head{N\textsubscript{pro}}[person]]
		[\Mod{Clause\textsubscript{rel}}
			[\Node{Prenucleus}{NP\textsubscript{x}}[who, roof]]
			[\Head{Clause\textsubscript{rel}}
				[\Subj{GAP\textsubscript{x}}[--]]
				[\Head{VP}[works, roof]]
			]
		]
	]
]
\end{forest}

Note that, as stipulated in \cref{sec:gapping-basics}, the gap is coindexed to just one overt antecedent---in this case, the relative pronoun \mention{who}. The pronoun's antecedent (\mention{person}) is therefore not coindexed.

In \mention{that} relatives and bare relatives, however, there is no extranuclear constituent, and so the co-indexation is with the antecedent, as in \cref{ex:barerelative,ex:thatrelative}.

\ex. \label{ex:barerelative} Bare relative:\\
    \begin{forest}
where n children=0{
    font=\itshape, 			
    tier=word          			
  }{
  },
[NP
	[\Det{DP}[a, roof]]
	[\Head{Nom}
		[\Head{N\textsubscript{x}}[person]]
		[\Mod{Clause\textsubscript{rel}}
			[\Subj{NP}[you, roof]]
			[\Head{VP}
			    [\Head{V}[know, roof]]
			    [\Obj{GAP\textsubscript{x}}[--]]
			]
		]
	]
]
\end{forest}

\needspace{5em}
\ex. \label{ex:thatrelative} \mention{That} relative:\\
    \begin{forest}
where n children=0{
    font=\itshape, 			
    tier=word          			
  }{
  },
[NP
	[\Det{DP}[a, roof]]
	[\Head{Nom}
		[\Head{N\textsubscript{x}}[person]]
		[\Mod{Clause\textsubscript{rel}}
		    [\Node{Marker}{Sdr}[that]]
		    [\Head{\Clauserel}
    			[\Subj{NP}[you, roof]]
    			[\Head{VP}
    			    [\Head{V}[know, roof]]
    			    [\Obj{GAP\textsubscript{x}}[--]]
    			]
    		]
		]
	]
]
\end{forest}


See also fused relatives, \cref{sec:head-prenuc}.

\subsection{Open interrogative clauses}
Every open interrogative clause has a gap co-indexed to an interrogative phrase as in (\ref{ex:InvertedInt}) unless the there is no inversion, in which case the interrogative phrase is in situ as in \mention{\underline{Who} did that?} or \mention{You did \underline{what} to him?}. This contrasts with relative clauses in which there is always a gap, even in subject function.

\ex. \label{ex:InvertedInt} 
    \begin{forest}
where n children=0{
    font=\itshape, 			
    tier=word          			
  }{
  },
[Clause
	[\Node{Prenucleus}{NP\fbox{\textsubscript{x}}}[who, roof]]
	[\Head{Clause}
		[\Node{Prenucleus}{V\textsubscript{aux}\fbox{\textsubscript{y}}}[did]]
		[\Head{Clause}
			[\Subj{NP}[you, roof]]
			[\Head{VP}
				[\Head{GAP\fbox{\textsubscript{y}}}[--]]
				[\Comp{Clause}
					[\Head{VP}
						[\Head{V}[see]]
						[\Obj{GAP\fbox{\textsubscript{x}}}[--]]
					]
				]
			]
		]
	]
]
\end{forest}

\subsection{Hollow clauses}\label{sec:hollow}
Every hollow clause has a gap co-indexed to an antecedent.

\ex. \label{ex:HollowCl} 
    \begin{forest}
where n children=0{
    font=\itshape, 			
    tier=word          			
  }{
  },
    [Clause
    	[\Subj{NP\textsubscript{x}}[the box, roof]]
    	[\Head{VP}
    		[\Head{\Vaux}[was]]
    		[\PredComp{AdjP}
    			[\Head{AdjP}
    				[\Mod{AdvP}[too, roof]]
    				[\Head{AdjP}[heavy, roof]]
    			]
    			[\Node{Comp\textsubscript{ind}}{Clause}
    				[\Head{VP}
    					[\Mk{Sdr}[to]]
    					[\Head{VP}
    						[\Head{V}[lift]]
    						[\Obj{GAP\textsubscript{x}}[--]]
    					]
    				]
    			]
    		]
    	]
    ]
\end{forest}

\ex. \label{ex:HollowClTough} 
    \begin{forest}
where n children=0{
    font=\itshape, 			
    tier=word          			
  }{
  },
    [NP
    	[\Det{DP}[a, roof]]
    	[\Head{Nom}
            [\Head{Nom\textsubscript{x}}
        		[\Mod{AdjP}[tough, roof]]
        		[\Head{N}[box]]
            ]
            [\Node{Comp\textsubscript{ind}}{Clause}
                [\Head{VP}
                    [\Mk{Sdr}[to]]
                    [\Head{VP}
                        [\Head{V}[lift]]
                        [\Obj{GAP\textsubscript{x}}[--]]
                    ]
    			]
    		]
    	]
    ]
\end{forest}

Note that in \cref{ex:HollowClTough}, the gap is coindexed with the \func{head} Nom \mention{tough box}, even though strictly speaking the adjective phrase is not understood as part of the gapped material.

\ex. \label{ex:HollowForPurpose} 
    \begin{forest}
where n children=0{
    font=\itshape, 			
    tier=word          			
  }{
  },
    [NP
    	[\Det{DP}[a, roof]]
    	[\Head{Nom}
            [\Head{N\textsubscript{x}}[jar]]
            [\Mod{PP}
                [\Head{P}[for]]
                [\Comp{Clause}
                    [\Head{VP}
                        [\Head{VP}
                            [\Head{V}[catching]]
                            [\Obj{NP}[fireflies, roof]]
                        ]
                        [\Mod{PP\textsubscript{strand}}
                            [\Head{P}[with]]
                            [\Obj{GAP\textsubscript{x}}[--]]
                        ]
        			]
        		]
            ]
    	]
    ]
\end{forest}

\section{Coordination}
A number of non-basic coordination constructions include gaps. For examples, see \cref{sec:non-basic}.

\section{Non-gapped constructions}
The following structures have at least a notional gap, but any such gaps are not annotated in CGELBank.

\subsection{Comparatives}

CGEL writes some examples of comparative clauses with a gap, as in \mention{It is as deep as} [\mention{it is \_\_ wide}], but we view this as heuristic and do not include any such gaps in tree structure. CGELBank does, however, mark such clauses with searchable but non-visible features.

\subsection{Subjects in infinitival and participial clauses}\label{sec:subjinfpart}
CGELBank does not mark a gap for a subject in infinitival clause or participial clause, whether or not there is control or raising. As a result, the gaps and co-indexation below will not appear in the tree structure, regardless of whether they are controlled as in \cref{ex:NoSubj} or not, as in \cref{ex:NoSubjNot}.

\ex. \label{ex:NoSubj}
    \a. \mention{I\textsubscript{x} hope \_\_\textsubscript{x} to see her.}
    \b. \mention{They asked Pat\textsubscript{x} \_\_\textsubscript{x} to help them.}
    
\ex. \label{ex:NoSubjNot}
    \a. \mention{That would mean \_\_ starting over.}
    \b. \mention{She\textsubscript{x} has an invitation \_\_\textsubscript{x} to attend.}

Though there is no subject, the VP is still the head of a clause, as in \cref{ex:NoSubjs}.

\ex. \label{ex:NoSubjs}
    \begin{forest}
    where n children=0{
        font=\itshape, 			
        tier=word          			
      }{
      },
    [Clause
    	[\Head{VP}
    		[\Head{V}[try]]
    		[\Comp{Clause}
    			[\Head{VP}
    				[\Head{V}[opening]]
    				[\Obj{NP}[the door,roof]]
    			]
    		]
    	]
    ]
\end{forest}

\subsection{Ellipsis} \label{sec:ellipsis}
Similar to the case in comparative clauses, we view CGEL's ``gaps'' indicating ellipsis to be heuristic. For treatment of trees with ellipsis, see \cref{sec:MiscEllipsis}.

\section{Co-indexing without gaps}
CGELBank does not co-index other anaphora, including displaced or extraposed constituents (but see \cref{sec:pre}).

\chapter{Coordination}\label{ch:coord}
\section{Introduction}

CGELBank generally follows CGEL in its analyses, but, consistent with SIEG2, we are explicit in taking coordinates to be headed phrases and markers to be dependents therein. This parallels CGEL's treatment of markers in subordination (9, p.~954).

Also, CGEL labels coordinations with the category of \func{coordinates} (e.g., NP-coordination). In contrast, CGELBank simply labels all coordinations as ``Coordination''. The nature of the coordination is established by the categories of the coordinates (or, nested coordinates, in the case of layered coordination; \cref{sec:layeredcoord}).

Here, we show how this difference in analysis affects the trees in CGELBank, while, at the same time, setting out our analysis of a number of points about which CGEL is inexplicit.

We also treat supplementation here, not because we disagree with CGEL's analysis, but because supplements, like \func{coordinate} phrases, can have a coordinator as a \func{marker} and because, for simplicity, we deviate from CGEL's style of indicating \func{supplements} with an arrow pointing to the anchor in the tree (see 12 on p.~1354).

\section{Basic coordination}\label{sec:basiccoord}

The most basic cases of coordination have two phrasal \func{coordinates} with a coordinator functioning as a \func{marker} in the second \func{coordinate} as in (\ref{ex:mostbasic}).

\ex. \label{ex:mostbasic} 
    \begin{forest}
    where n children=0{
        font=\itshape, 			
        tier=word          			
      }{
      },
    [Clause
        [\Subj{NP}[I, roof]]
        [\Head{VP}
            [\Head{V}[like]]
            [\Obj{Coordination}
        	    [\Node{Coordinate}{NP}[cats, roof]]
        	    [\Node{Coordinate}{NP}
        		    [\Mk{Coordinator}[and]]	[\Head{NP}[dogs, roof]]
            	]
            ]
        ]
    ]
    \end{forest}

NP-coordination is preferred over Nom-coordination if both readings are equally plausible. An example that requires the latter (with a shared determiner):

\ex. \label{ex:mostbasic2} 
    \begin{forest}
    where n children=0{
        font=\itshape, 			
        tier=word          			
      }{
      },
    [NP
        [\Det{DP}[a, roof]]
        [\Head{Coordination}
            [\Node{Coordinate}{Nom}[cat, roof]]
            [\Node{Coordinate}{Nom}
                [\Mk{Coordinator}[or]]	[\Head{Nom}[dog, roof]]
            ]
        ]
    ]
    \end{forest}

A Coordination should never be the head of a unary-branching constituent. Thus, a clause with a shared subject may be headed by a VP-coordination, but a clause with no subject or other external dependents cannot be headed by a VP-coordination. Instead, the coordination should be of two clauses.\footnote{E.g., \mention{They will \emph{[}\uline{eat} \,\uline{or perhaps drink}\emph{]}}: the bracketed portion is a Coordination with two underlined Clause constituents as \func{coordinate}s.}

\subsection{Asyndetic coordination}\label{sec:asyndetic}

Asyndetic coordination, as in (\ref{ex:asyndeton}), lacks a \func{marker} such as \mention{and}. Most of what is presented here will deal explicitly with syndetic coordination, but will almost always apply equally to asyndetic coordination, except, of course, for discussion of \func{markers}. 

\ex. \label{ex:asyndeton} 
    \begin{forest}
    where n children=0{
        font=\itshape, 			
        tier=word          			
      }{
      },
    [Clause
    	[\Head{VP}
    	    [\Head{V}[bring]]
    		[\Obj{Coordination}
    			[\Node{Coordinate}{NP}[games, roof]]	[\Node{Coordinate}{NP}[stories, roof]]
    			[\Node{Coordinate}{NP}[songs, roof]]
    		]
    	]
    ]
    \end{forest}

\subsection{Coordination of lexemes}\label{sec:lexcoord}
CGELBank treats apparent coordination of lexemes as coordination of phrases, as in \cref{ex:lexemes6}. (Appendix~\ref{app:lexcoord} presents an in-depth comparison of alternatives to justify this approach.) The one exception to this is discussed in \cref{sec:whether}.

\ex.\label{ex:lexemes6}
    \begin{forest}
    where n children=0{
        font=\itshape, 			
        tier=word          			
      }{
      },
    [VP
		[\Head{Coordination}
			[\Node{Coordinate}{\fbox{VP}}
			    [\Head{V\textsubscript{aux}}[can]]]
			[\Node{Coordinate}{\fbox{VP}}
				[\Mk{Coordinator}[and]]	[\Head{\fbox{VP}}
				    [\Head{V\textsubscript{aux}}[will]]]
			]
		]
		[\Comp{Clause}[try, roof]]
	]
    \end{forest}
    
\ex.\label{ex:lexemes7}
    \begin{forest}
    where n children=0{
        font=\itshape, 			
        tier=word          			
      }{
      },
    [NP
    	[\Head{Nom}
    		[\Head{Coordination}
    			[\Node{Coordinate}{\fbox{Nom}}
    			    [\Head{N}[development]]]
    			[\Node{Coordinate}{\fbox{Nom}}
    				[\Mk{Coordinator}[and]]	[\Head{\fbox{Nom}}
    				    [\Head{N}[implementation]]]
    			]
    		]
    		[\Comp{PP}[of policy]]
    	]
    ]
    \end{forest}
    
\ex.\label{ex:lexemes8}
    \begin{forest}
    where n children=0{
        font=\itshape, 			
        tier=word          			
      }{
      },
    [AdjP
    	[\Mod{AdvP}[very, roof]]
		[\Head{Coordination}
			[\Node{Coordinate}{\fbox{AdjP}}
			    [\Head{Adj}[friendly]]]
			[\Node{Coordinate}{\fbox{AdjP}}
				[\Mk{Coordinator}[and]]	[\Head{\fbox{AdjP}}
				    [\Head{Adj}[helpful]]]
			]
		]
    ]
    \end{forest}

\needspace{3cm}
\subsection{Coordinated-head NP with a shared post-head dependent}\label{sec:shared-post-head}

In NP structure with coordination in the head, there may be a shared post-head PP complement, as in \cref{ex:coord-heads-posthead-comp}:

\ex. \label{ex:coord-heads-posthead-comp}
    \a.\label{ex:coord-comp-simple} [$_{\text{Nom}}$ [photos and video recordings] [of FDR]]
    \b.\label{ex:coord-comp-delayed} [$_{\text{NP}}$ [$_{\text{Coordination}}$ [$_{\text{NP}}$ all [$_{\text{Nom}}$ photos --$_i$]] and [$_{\text{NP}}$ most [$_{\text{Nom}}$ video recordings --$_i$]]] [of~FDR]$_i$]
    \z.

The PP complement is analyzed as an internal dependent within the Nom, just as it would be absent coordination.
In \cref{ex:coord-comp-simple}, this is achieved with coordination of two Nom constituents, like \cref{ex:lexemes7}.
If there is a coordination that has NPs as its coordinates, as with \cref{ex:coord-comp-delayed}, the whole structure is analyzed as delayed right constituent coordination (\cref{sec:delayed}).

PP modifiers (e.g., replacing \mention{of FDR} in \cref{ex:coord-comp-simple,ex:coord-comp-delayed} with \mention{in the library}) follow the same structure.


When a coordination of NPs is relativized and there is no delayed right constituent coordination interpretation, we are forced to make the relative clause an external (peripheral) modifier:

\ex. \label{ex:external-rc} 
    \begin{forest}
    where n children=0{
        font=\itshape, 			
        tier=word          			
      }{
      },
    [NP
        [\Head{Coordination}
            [\Node{Coordinate}{NP}[the cat, roof]]
            [\Node{Coordinate}{NP}[and the mouse, roof]]
        ]
        [\Mod{Clause\textsubscript{rel}}
            [\Node{Prenucleus}{NP\textsubscript{x}}[who, roof]]
            [\Head{Clause\textsubscript{rel}}[GAP\textsubscript{x} live in the basement, roof]
            ]
        ]
    ]
    \end{forest}

\subsection{Root-branched coordinators and coordinates} \label{sec:root}
Here we consider cases in which there is a sentence-initial coordinator or where the root is a coordination. (The related case of a \func{supplement} marked with a coordinator will be dealt with in \cref{sec:supplments}.) 

Where a sentence starts with a coordinator, we give that sentence its usual constituent label, so that (\ref{ex:rootclause}) \mention{And so it begins} is a clause, and (\ref{ex:rootNP}) \mention{But not that} is an NP.

\ex. \label{ex:rootclause} 
    \begin{forest}
    where n children=0{
        font=\itshape, 			
        tier=word          			
      }{
      },
    [Clause
    	[\Mk{Coordinator}[and]]
    	[\Head{Clause}[so it begins, roof]]
    ]
    \end{forest}
    
\ex. \label{ex:rootNP} 
    \begin{forest}
    where n children=0{
        font=\itshape, 			
        tier=word          			
      }{
      },
    [NP
    	[\Mk{Coordinator}[but]]
    	[\Head{NP}[not that, roof]]
    ]
    \end{forest}

\noindent When the sentence consists of two or more coordinates, the root is a coordination, not a clause or other XP, as in (\ref{ex:rootCoordination}).

\ex. \label{ex:rootCoordination} 
    \begin{forest}
    where n children=0{
        font=\itshape, 			
        tier=word          			
      }{
      },
    [Coordination
    	[\Node{Coordinate}{Clause}[stay, roof]]
    	[\Node{Coordinate}{Clause}
    		[\Mk{Coordinator}[and]]	[\Head{Clause}[have some coffee, roof]]
    	]
    ]
    \end{forest}

\subsection{Layered coordination}\label{sec:layeredcoord}



\ex. \label{ex:layered}
    \begin{forest}
    where n children=0{
        font=\itshape, 			
        tier=word          			
      }{
      },
    [Coordination
    	[\Node{Coordinate}{Coordination}[small and quiet, roof]]
    	[\Node{Coordinate}{Coordination}
    		[\Mk{Coordinator}[but]]
    		[\Head{Coordination}
    			[\Node{Coordinate}{AdjP}[artful, roof]]
    			[\Node{Coordinate}{AdjP}
    				[\Mk{Coordinator}[and]]
    				[\Head{AdjP}[enterprising, roof]]
    			]
    		]
    	]
    ]
\end{forest}



\subsection{Correlative coordination and marker category}\label{sec:correlative-coord}


A DP may function as \func{marker} in the first coordinate of a coordination in correlative coordination. (See also CGEL's [45] on p.~1308 for an example of a gapped marker.)

\ex. \label{ex:correlative}
    \begin{forest}
    where n children=0{
        font=\itshape, 			
        tier=word          			
      }{
      },
    [Coordination
    	[\Node{Coordinate}{AdjP}
    		[\Mk{DP}[neither, roof]]
    		[\Head{AdjP}[artful, roof]]
    	]
    	[\Node{Coordinate}{AdjP}
    		[\Mk{Coordinator}[nor]]
    		[\Head{AdjP}[enterprising, roof]]
    	]
    ]
\end{forest}

A subordinator may also appear in \func{marker} function, so that it's possible to have strings of two \func{markers} like \mention{either to} and \mention{or to} in (\ref{ex:correlative2}).

\ex. \label{ex:correlative2}
    \begin{forest}
    where n children=0{
        font=\itshape, 			
        tier=word          			
      }{
      },
    [Coordination
    	[\Node{Coordinate}{VP}
    		[\Mk{DP}[either, roof]]
    		[\Head{VP}
    		    [\Mk{Sdr}[to]]
    		    [\Head{VP}[live, roof]]
    		]
    	]
    	[\Node{Coordinate}{VP}
    		[\Mk{Coordinator}[or]]
    		[\Head{VP}
    		    [\Mk{Sdr}[to]]
    		    [\Head{VP}[let live, roof]]
    		]
    	]
    ]
    \end{forest}

\subsection{Expansion of coordinates by modifiers}

The  structure in CGELBank differs from CGEL's (see CGEL's [9] on p.~1278). Here, the \mbox{AdvPs} in an NP like \mention{the guests and indeed his family too} from (\ref{ex:expansion}) are analyzed as \func{peripheral modifiers} or possibly, in the case of \textit{indeed}, as \func{supplements}. 

\ex. \label{ex:expansion}
    \begin{forest}
    where n children=0{
        font=\itshape, 			
        tier=word          			
      }{
      },
    [Coordination
    	[\Node{Coordinate}{NP}[the guests, roof]]
    	[\Node{Coordinate}{NP}
    		[\Mk{Coordinator}[and]]
    		[\Sup{AdvP}[indeed, roof]]
    		[\Head{NP}
    			[\Head{NP}[his family, roof]]
    			[\Mod{AdvP}[too, roof]]
    		]
    	]
    ]
    \end{forest}

\section{Non-basic coordination} \label{sec:non-basic}

\subsection{Right nonce-constituent coordination}

CGEL gives no function to a right nonce-constituent coordination (p.~1342), but to avoid this functionless node, CGELBank uses the + operator to create a nonce function, as in  (\ref{ex:rightnonce}).

\ex. \label{ex:rightnonce}
    \begin{forest}
    where n children=0{
        font=\itshape, 			
        tier=word          			
      }{
      },
    [Clause
        [\Subj{NP}[Mo, roof]]
        [\Head{VP}
            [\Head{V}[gave]]
            [\Node{Obj\textsubscript{ind} + Obj\textsubscript{dir} + Mod}{Coordination}
               [\Node{Coordinate}{NP + NP + PP}
                    [\Node{Obj\textsubscript{ind}}{NP}[me, roof]]
                    [\Node{Obj\textsubscript{dir}}{NP}[one, roof]]
        	        [\Mod{PP}[before, roof]]
                ]
        		[\Node{Coordinate}{NP + NP + PP}
        			[\Mk{Coordinator}[and]]
            		[\Head{NP + NP + PP}
                		[\Node{Obj\textsubscript{ind}}{NP}[Jo, roof]]
                        [\Node{Obj\textsubscript{dir}}{NP}[two, roof]]
                		[\Mod{PP}[after, roof]]
            		]
        		]
        	]
        ]
    ]
    \end{forest}

\subsection{Gapped coordination}

CGEL calls examples like \mention{Kim is\textsubscript{x} an engineer and Pat \_\_\textsubscript{x} a barrister} ``gapped coordinations'' and includes gaps in the phrase structure. In contrast, CGELBank treats them mostly like right nonce-constituent coordinations: 

\ex. \label{ex:gapped1}
    \begin{forest}
    where n children=0{
        font=\itshape, 			
        tier=word          			
      }{
      },
    [Coordination
    	[\Node{Coordinate}{Clause}
    		[\Subj{NP}[Kim, roof]]
    		[\Head{VP}[is an engineer, roof]]
    	]
    	[\Node{Coordinate}{NP + NP}
    		[\Mk{Coordinator}[and]]
    		[\Head{NP + NP}
    			[\Subj{NP}[Pat, roof]]
    			[\PredComp{NP}[a barrister, roof]]
    		]
    	]
    ]
    \end{forest}

\noindent This also applies to more complex cases in which the ``gap'' is not a constituent but a string like \mention{wanted him to marry}

\ex. \label{ex:gapped2}
    \begin{forest}
    where n children=0{
        font=\itshape, 			
        tier=word          			
      }{
      },
    [Coordination
    	[\Node{Coordinate}{Clause}
    		[\Subj{NP}[his father, roof]]
    		[\Head{VP}[wanted him to marry Sue, roof]]
    	]
    	[\Node{Coordinate}{NP + NP}
    		[\Mk{Coordinator}[but]]
    		[\Head{NP + NP}
    			[\Subj{NP}[his mother, roof]]
    			[\Obj{NP}[Louise, roof]]
    		]
    	]
    ]
    \end{forest}

An interesting case occurs in coordinated verbless \func{complements} of \mention{with}, as in (\ref{ex:gapped3}). This example also illustrates that in gapped coordination the function of the coordination, if it is not the root, will usually be a typical function like \func{Comp}, as opposed to the kind of nonce function found in right nonce-constituent coordination.

\ex. \label{ex:gapped3}
    \begin{forest}
    where n children=0{
        font=\itshape, 			
        tier=word          			
      }{
      },
    [PP
    	[\Head{P}[with]]
    	[\Comp{Coordination}
    		[\Node{Coordinate}{NP + AdjP}
    			[\Subj{NP}[Jill, roof]]
    			[\PredComp{AdjP}[intent on staying, roof]]
    		]
    		[\Node{Coordinate}{NP + PP}
    			[\Mk{Coordinator}[and]]
    			[\Head{NP + PP}
    				[\Subj{NP}[Pat, roof]]
    				[\Comp{NP}[on leaving, roof]]
    			]
    		]
    	]
    ]
    \end{forest}

The coordinates in this construction can have more than two constituents, as in (\ref{ex:gapped4}), which is the same string as (\ref{ex:rightnonce}) but semantically and structurally quite different. 

\ex. \label{ex:gapped4}
    \begin{forest}
    where n children=0{
        font=\itshape, 			
        tier=word          			
      }{
      },
    [Coordination
        [\Node{Coordinate}{Clause}
            [\Subj{NP}[Mo, roof]]
        	[\Head{VP}
        	    [\Head{VP}
            	   [\Head{V}[gave]]
            	   [\Node{Obj\textsubscript{ind}}{NP}[me, roof]]
            	   [\Node{Obj\textsubscript{dir}}{NP}[one, roof]]
            	]
    	        [\Mod{PP}[before, roof]]
    		]
    	]
		[\Node{Coordinate}{NP + NP + PP}
			[\Mk{Coordinator}[and]]
    		[\Head{NP + NP + PP}
        		[\Node{Subj}{NP}[Jo, roof]]
                [\Node{Obj\textsubscript{dir}}{NP}[two, roof]]
        		[\Mod{PP}[after, roof]]
    		]
		]
    ]
    \end{forest}

\subsection{Delayed right constituent coordination} \label{sec:delayed}

CGELBank treat these constructions as having a gap and a postnucleus.

\ex. \label{ex:delayed} 
    \begin{forest}
    where n children=0{
        font=\itshape, 			
        tier=word          			
      }{
      },
    [Clause
    	[\Subj{NP}[I, roof]]
    	[\Head{VP}
    		[\Head{Coordination}
    			[\Node{Coordinate}{VP}
    			    [\Head{V}[saw]]
    			    [\Obj{GAP\textsubscript{x}}[--]]
    			]
    	        [\Node{Coordinate}{VP}
    			    [\Mk{Coordinator}[but]]
        			[\Head{VP}
                        [\Head{V\textsubscript{aux}}[didn't]]
                        [\Comp{Clause}
                            [\Head{VP}
                                [\Head{V}[meet]]
                                [\Obj{GAP\textsubscript{x}}[--]]
                            ]
                        ]
        			]
        		]
    		]
    		[\Node{Postnucleus}{NP\textsubscript{x}}[her, roof]
    		]
    	]
    ]
    \end{forest}

CGEL takes no clear position on cases like \mention{He’s as old as or older than me}, where there is no prosodic break before the final element, and where it can be an unstressed personal pronoun (p.~1345). CGELBank takes the same approach as above, as illustrated in (\ref{ex:delayed2}).

\ex. \label{ex:delayed2} 
    \begin{forest}
    where n children=0{
        font=\itshape, 			
        tier=word          			
      }{
      },
        [AdjP
            [\Head{Coordination}
                [\Node{Coordinate}{AdjP}
                    [\Head{AdjP}[as old, roof]]
                    [\Node{Comp\textsubscript{ind}}{PP}
                        [\Head{P}[as]]
                        [\Comp{GAP\textsubscript{x}}[--]]
                    ]
                ]
    	        [\Node{Coordinate}{AdjP}
    			    [\Mk{Coordinator}[or]]
        			[\Head{AdjP}
                        [\Head{Adj}[older]]
                        [\Comp{PP}
                            [\Head{P}[than]]
                            [\Comp{GAP\textsubscript{x}}[--]]
                        ]
                    ]
                ]
            ]
            [\Node{Postnucleus}{NP\textsubscript{x}}[me, roof]]
        ]
    \end{forest}


\subsection{End-attachment coordination}
\subsubsection{Postposing of coordinate}

\ex. \label{ex:end} 
    \begin{forest}
    where n children=0{
        font=\itshape, 			
        tier=word          			
      }{
      },
    [Clause
    	[\Subj{NP}[I, roof]]
    	[\Head{VP}
    		[\Head{VP}
    			[\Head{V}[made]]
    			[\Obj{Coordination}
    				[\Node{Coordinate}{NP}[this one, roof]]
    				[\Node{Coordinate}{GAP\textsubscript{x}}[--]]
    			]
    			[\PredComp{AdjP}[too sweet, roof]]
    		]
    		[\Node{Postnucleus}{NP\textsubscript{x}}[but not that one, roof]]
    	]
    ]
    \end{forest}

\subsubsection{Addition of a new element}
This construction is not, in fact, a coordination. Instead, the coordinator-initial constituent that looks like a final \func{coordinate} is a \func{supplement} (see also \cref{sec:Supplements} and \cref{sec:supplments}).

\ex. \label{ex:end2} 
    \begin{forest}
    where n children=0{
        font=\itshape, 			
        tier=word          			
      }{
      },
    [Clause
    	[\Subj{NP}[I, roof]]
		[\Head{VP}[knew her, roof]]
    	[\Sup{AdvP}[but not well, roof]]
    ]
    \end{forest}

As this is not a \func{coordinate}, however, the Coordinator is allowed to be sister to a lexical \func{head}, subject to the usual preference for binary branching:

\ex. \label{ex:end3} 
    \begin{forest}
    where n children=0{
        font=\itshape, 			
        tier=word          			
      }{
      },
    [Clause
    	[\Subj{NP}[I, roof]]
		[\Head{VP}[met her, roof]]
    	[\Sup{AdvP}
            [\Mk{Coordinator}[but]] 
            [\Head{Adv}[briefly]]]
    ]
    \end{forest}

\section{Individual constructions with coordinators}
\subsection{\mention{Not only} X \mention{but} Y}

At the beginning of a \mention{but}-coordination, \mention{not} (\mention{only/just/even\dots}) bears parallels to a marker of correlative coordination as in \cref{sec:correlative-coord} (\mention{either\dots or}, \mention{both\dots and}, etc.), but is better analyzed as a modifier (pp.~1313--1314).

\ex. \label{ex:notonlybut-simple} 
    \begin{forest}
    where n children=0{
        font=\itshape, 			
        tier=word          			
      }{
      },
    [Coordination
    	[\Node{Coordinate}{Clause}
            [\Subj{NP}[he, roof]]
            [\Head{VP}
                [\Head{V\textsubscript{aux}}[was]]
                [\PredComp{AdjP}
                    [\Mod{AdvP}[not only, roof]]
                    [\Head{Adj}[right]]
                ]
            ]
    	]
    	[\Node{Coordinate}{Clause}
    		[\Mk{Coordinator}[but]]
    		[\Head{Clause}
    				[\Subj{NP}[he, roof]]
    				[\Head{VP}
    					[\Head{V\textsubscript{aux}}[was]]
    					[\PredComp{AdjP}[prescient, roof]]
    				]
    			]
    		]
    	]
    ]
    \end{forest}

The construction exemplified in \cref{ex:notonlybut-inversion} is a special case of inversion (see \cref{sec:AdjInv}):

\ex. \label{ex:notonlybut-inversion} 
    \begin{forest}
    where n children=0{
        font=\itshape, 			
        tier=word          			
      }{
      },
    [Coordination
    	[\Node{Coordinate}{Clause}
    		[\Node{Prenucleus}{AdvP\fbox{\textsubscript{x}}}[not only, roof]]
    		[\Head{Clause}
    			[\Node{Prenucleus}{V\textsubscript{aux}\fbox{\textsubscript{y}}}[was]]
    			[\Head{Clause}
    				[\Subj{NP}[he, roof]]
    				[\Head{VP}
    				    [\Head{VP}
    					    [\Head{GAP\fbox{\textsubscript{y}}}[--]]
    					]
    					[\PredComp{AdjP}
    					    [\Mod{GAP\fbox{\textsubscript{x}}}[--]]
    					    [\Head{Adj}[right]]
    					]
    				]
    			]
    		]
    	]
    	[\Node{Coordinate}{Clause}
    		[\Mk{Coordinator}[but]]
    		[\Head{Clause}
    				[\Subj{NP}[he, roof]]
    				[\Head{VP}
    					[\Head{V\textsubscript{aux}}[was]]
    					[\PredComp{AdjP}[prescient, roof]]
    				]
    			]
    		]
    	]
    ]
    \end{forest}

If \mention{but} were omitted from the sentence, it would be analyzed as asyndetic coordination (\cref{sec:asyndetic}, but see CGEL p.~1314 noting that this analysis is debatable).


\subsection{\mention{Whether}}\label{sec:whether}
\mention{Whether} is an unusual subordinator in what it coordinates with. The expressions \mention{whether or not} and \mention{whether or no} are complex subordinators (see \cref{sec:complex}) and not coordinations. Nevertheless, \mention{whether} can coordinate with interrogative phrases (e.g., \textit{whether and why we would go}), including PPs as in \mention{whether and to what extent}. Examples like those in \cref{ex:Whetherand} are the only cases where a coordinate is a lexeme.

\ex. \label{ex:Whetherand} 
    \begin{forest}
    where n children=0{
        font=\itshape, 			
        tier=word          			
      }{
      },
        [Clause
        	[\Node{Marker+Prenucleus}{Coordination}
        		[\Node{Coordinate}{Sdr}[whether]]
        		[\Node{Coordinate}{PP}
        			[\Mk{Coordinator}[and]]
        			[\Head{PP}[to what extent, roof]]
        		]
        	]
        	[\Head{Clause}[it works, roof]]
        ]
    \end{forest}

\subsection{\mention{Etc.}}\label{sec:Etc}
\mention{Etc.}~is literally ``and others'', so we take it to be a coordinator in fused \func{marker-head} function (see \cref{sec:marker-head}).

\ex. \label{ex:Etc} 
    \begin{forest}
    where n children=0{
        font=\itshape, 			
        tier=word          			
      }{
      },
    [Coordination
    	[\Node{Coordinate}{NP}[books, roof]]
    	[\Node{Coordinate}{NP}[pencils, roof]]
    	[\Node{Coordinate}{NP},s sep=-2em
    		[\phantom{X}\hspace*{-4em},tier=dh]
    		[\small\textsf{Marker-Head:}\\Coordinator,no edge,tier=dh
    			[\textit{etc.}]
    		]
    		[\small\textsf{Head:}\\Nom
    			[\hspace*{-4em}\phantom{X},tier=dh]
    		]
    	]
    ]
    \end{forest}

\subsection{x to y ranges}\label{sec:hyphenranges} 

We take cases like \mention{5:00--8:00} in phrases like \mention{between 5:00--8:00} or on their own to be coordinations, with the hyphen \mention{-} or \mention{--} or being a coordinator read ``and'' or ``to'' or having no phonological realization.\footnote{These should not be confused with coordinative compounds such as \mention{Austria-Hungary}, \mention{Hewlett-Pakard}, or \mention{murder-suicide}, which, like all compounds, are individual lexical items.} The slash \mention{/} is a possible alternative which can also mean ``or'' (p.~1764). There are also instances with just a space, such as \mention{this Earth Moon highway}. We take these to be asyndetic coordinations.

\ex. \label{ex:hyphenranges} 
    \a. \mention{I work \underline{Mon--Fri}/\mention{\underline{Mon to Fri}.}}
    \b. \mention{John Nash (\underline{1928--2015})}
    \b. \mention{a \underline{French--English}}/\mention{\underline{French/English} dictionary}
    \b. \mention{the \underline{May/June} period}
    \b. \mention{a \underline{parent--teacher} meeting}
    \b. \mention{pp. \underline{23--64}}

In contrast, in constructions like \mention{We went from Toronto to Georgetown}, the PPs \mention{from Toronto} and \mention{to Georgetown} are a source complement and a goal complement respectively (p.~258). And the structure of cases like \mention{\underline{From Toronto to Georgetown}} / \mention{\underline{To Georgetown from Toronto} is about eight and a half hours} has the second PP as a complement in the first (p.~433).


\subsection{Supplements marked by a coordinator}\label{sec:supplments} 

CGEL notes that \func{supplements} may be marked by a coordinator (p.~1361).\footnote{A sentence consisting solely of a constituent marked by a coordinator is not a supplement (see \cref{sec:root}).} An example showing the representation in CGELBank is given in (\ref{ex:mksupp}). See \cref{sec:Supplements} for the structure of supplements.

\ex. \label{ex:mksupp} 
    \begin{forest}
    where n children=0{
        font=\itshape, 			
        tier=word          			
      }{
      },
    [Clause
        [\Subj{NP}[I, roof]]
        [\Head{VP}
            [\Head{V}[said]]
            [\Sup{Clause}
                [\Mk{Coordinator}[and]]
                [\Head{Clause}[I still believe it, roof]]
            ]
            [\Comp{Clause}[that you should try, roof]]
        ]
    ]
    \end{forest}


\chapter{Miscellaneous: Ellipsis, errors, punctuation, and future goals}
\section{Ellipsis} \label{sec:MiscEllipsis}
As noted in \cref{sec:ellipsis}, we do not use ``gaps'' to show ellipted material. The following examples are provided to illustrate where a gap would \textbf{not} appear in CGELBank.

\subsection{Elliptical stranding (Post-auxiliary ellipsis, including subordinator \mention{to})}\label{sec:ellipstrand}
Elliptical stranding includes no gaps. CGEL sometimes marks gaps in examples with post-auxiliary ellipsis such as \mention{I’ll \underline{help you}\textsubscript{x} if I can \ldots\textsubscript{x}.} In such a case, the contents of the gap may be inferred from elements in a higher clause, but it might have to be inferred from the context (e.g., \mention{Do you think we should \ldots?}). This means it's not always possible to have a co-index. We, therefore, deal with this by assuming that the VPs contain only the auxiliary verb and no complement. 

In the case of \mention{to}, this is not possible.\footnote{We would prefer to analyze \mention{to} as a highly defective auxiliary \citep{Levine2012}, but we decided to follow CGEL instead.} CGEL gives no guidance on the tree structure here. We, therefore, use a fused \func{Marker-Head}. (See \cref{sec:Etc} for an example with \mention{etc.})


\subsection{Ellipsis of postmodifiers}
[\mention{An article \underline{on this topic}\textsubscript{x}}] \mention{is more likely to be accepted than }[\mention{a book} \ldots\textsubscript{x}].

\subsection{In response to questions}
A: \mention{Whose \underline{father}\textsubscript{x} is on duty today?} B: \mention{Kim’s \ldots\textsubscript{x}.}

\subsection{With \mention{let's}}
A: \mention{Let's go\textsubscript{x}.} B: \mention{Yes, let's \ldots\textsubscript{x}.}

\subsection{Reduced interrogative clauses}
\mention{He made some mistakes\textsubscript{x}, though I don’t know how many \ldots\textsubscript{x}.}

\subsection{Subclausal coordination}
\noindent\mention{\underline{There is a copy}\textsubscript{x}} [\mention{on the desk and \ldots\textsubscript{x} in the top drawer}].

\noindent\mention{We’ll be in\textsubscript{x} Paris for a week and \ldots\textsubscript{x} Bonn for three days.} 
(see \cref{ch:coord} and CGEL's fn~65 on p.~1343)

\subsection{Ellipsis of complement of lexical verbs and adjectives}
A: \mention{Hawaii\textsubscript{x} would be nice, but I can't \underline{go \ldots\textsubscript{x}}\textsubscript{y}.} B: \mention{But you promised \ldots\textsubscript{y}.}

\subsection{Ellipsis of subject}
In CGELBank, such sentences are treated as VPs.

\subsubsection{Ellipsis of personal pronoun subject}
\mention{\ldots Doesn't matter.} (\mention{it})

\subsubsection{Ellipsis of subject pronoun + auxiliary}
\mention{\ldots Never seen anything like it!} (\mention{I have})

\subsubsection{Ellipsis in closed interrogatives of auxiliary or auxiliary + subject pronoun}
\mention{\ldots Feeling any better?} (\mention{are you})

\subsection{Determiner in NP structure}
\mention{\ldots Trouble is, we have to be there by six.} (\mention{the})

\subsection{Radical ellipsis in open interrogatives}
Such sentences are usually phrases. For example, the following is an AdvP.

A: \mention{I'm leaving.} B: \mention{Why \ldots.} (\mention{are you leaving})

\subsection{Subordinate clauses}
Cases like this are treated as phrases. For example, the complement of \mention{wonder} in the following is an AdvP. 

A: \mention{\underline{They got in without a key}\textsubscript{x}.} B: \mention{I wonder how \ldots\textsubscript{x}.}

\subsection{Radical ellipsis in declarative responses}
Again, such sentences are phrases. Here, \mention{yesterday} is an NP.

A: \mention{When did she get home?} B: \mention{\ldots yesterday.} (\mention{she got home})

\section{Errors}\label{sec:errors}
Where there is a clear error, both the original form and the corrected form are included in the tree: 
the original form with a \texttt{:t} value (for ``token'' or ``terminal''), 
and the corrected form with a \texttt{:correct} value. 
The lemma should reflect the correct version of the word, 
and must be indicated explicitly if it differs from the \texttt{:correct} value.

In \cref{ex:spelling}, for instance, the \textbf{misspelling} \mention{out} has been corrected to \mention{our}, and its lemma is \mention{we}.
In the graphical display, the corrected form will be included in parentheses after the original form.\footnote{The overarching description of the raw data format appears in Appendix~\ref{ch:format}.}

\ex.\label{ex:spelling} \texttt{(N\_pro :t "out" :correct "our" :l "we")}

\subsection{Omissions}
Consider \mention{*They are preparing my older son for kindergarten and looks forward to seeing his teacher and friends everyday.} Presumably, it is the son looking forward to things, so the subject of \mention{looks} should probably be \mention{he}.
A pronoun token is thus inserted for the missing word; it has no original token (\texttt{:t} value), so it is possible to recognize it as an insertion.

\ex.\label{ex:omission} \texttt{(N\_pro :correct "he")}

\subsection{Extra words}
The example \mention{go to the room 401} illustrates an apparent grammatical error: \mention{the} should be omitted in this context, though it is close to its ordinary use as a determinative in \func{determiner} function. We retain the word in the tree with an empty string value for \texttt{:correct}:

\ex.\label{ex:strike} \texttt{(D :t "the" :correct "")}

We have not yet encountered any sentences with extra words due to speech repair or total incoherence. Such words might be deemed unparseable and removed from the tree.


\subsection{Non-standard morphology}
Where this seems to be a dialectal form, no correction is indicated. Where it appears to be an error, it's treated as a misspelling. For example \mention{*I am works hard} should presumably be \mention{I am working hard}.

\subsection{Punctuation}\label{sec:punc-enc}
Punctuation is considered to have little impact on tree structure (see \cref{sec:punctuation}) and is not included in the graphical trees in this document (except as part of a lexical item, like hyphens in compounds and apostrophes in genitive nouns).
In the raw trees, such extra-lexical punctuation tokens are included as \texttt{:p} values alongside lexical tokens:

\ex.\label{ex:puncts} \texttt{(V :p "(" :t "microwaved" :l "microwave" :xpos "VBN" :p "?" :p ")" :p~",")}

Note that there can be multiple \texttt{:p} tokens within a lexical node, and their order and position with respect to the lexical token is significant: \cref{ex:puncts} represents the string \texttt{(~microwaved~?~)~,}~with space-separated punctuation or \texttt{(microwaved?),}~without space-separated punctuation.

Punctuation tokens must be associated with a neighboring lexical node (not a gap). By convention, CGELBank (starting from version 1.1) places any sentence-initial punctuation sequences, and any punctuation sequences beginning with open-parentheses/brackets, with the subsequent lexical token. All other punctuations are grouped with the preceding lexical token.

CGELBank trees do not themselves include any notations for the presence or absence of spacing around punctuations (though the original sentence string is provided).
CGELBank does not indicate corrections for nonstandard use or omission of extra-lexical punctuation tokens.

\subsection{Syntax}
Apart from errors noted above, CGELBank does not annotate errors of syntax. For example, the verb \textit{said} does not license an indirect object. Nevertheless, in \mention{I said him the answer}, no correction would be noted.

\section{Punctuation and symbols} \label{sec:punctuation}
In CGELBank, punctuation is included in raw trees (as described in \cref{sec:punc-enc}) but omitted from graphical trees.
Syntactically, punctuation often has little impact except for marking the end of a sentence or marking supplements.

\subsection{Primary terminals}
A tree typically consists of a sentence, which ends with a primary terminal.

\subsection{Marking supplements}
Supplements are often set off by punctuation, as in \cref{ex:supPunc}.

\ex. \label{ex:supPunc}
    \a. \mention{An official\underline{, who spoke on the condition of anonymity,} said it was unlikely to succeed.}
    \b. \mention{Anderson and Der Khosla of the Bureau of Competition Policy \underline{(the ``Bureau'')} defined industrial policy as \ldots}
    \b. \mention{It was questionable at the time\underline{; however, it's now the consensus view}.}

\subsection{Hyphens}
See \cref{sec:hyphenranges}.

\subsection{Position of currency symbols}\label{sec:currency}
Currency symbols like \mention{\$} for \mention{dollar(s)} should be placed in the order pronounced, regardless of the original orthography. That is, the untokenized string \mention{\$300} would be tokenized and ordered in the tree as \mention{300~\$}, and analyzed with the same structure as \mention{300~dollars}. 
This is necessary as the quantity phrase may be complex (e.g., \mention{over \$300} analyzed as \mention{[over 300] dollars}).

\subsection{Emoji}
We treat emoji as interjections in \func{supplement} function.

\section{Future goals}\label{sec:future}

As additional data is encountered, additional clarifications to the guidelines are expected to become necessary.

Based on the data encountered already, we feel that a few areas merit further deliberation or enhancement.

\subsection{Potential points of revision}

\paragraph{Verbless clauses.} The structure of what CGEL terms verbless clauses (e.g., \mention{\uline{With the baby asleep}, we can go about our business}) is unclear. 
Is it appropriate to consider them a subtype of clause if there is no verb?
Further discussion is necessary.

\paragraph{External modifiers.} Within NPs, CGEL describes a distinction between internal and external modifiers. The external ones are outside the Nom, and thus structurally identifiable in NPs. But some of them---notably focusing modifiers---can also modify other kinds of phrases. Is a designated external modifier function warranted?

\paragraph{Medial modifiers.} CGEL speaks of a notion of \emph{core} complement in a VP. Objects, perhaps, are core whereas PP complements are not, as evidenced by adverbs' resistance to placement between verb and object. But it is not clear how uniform this pattern is. Does it warrant a revision to the branching rules within VPs?

\paragraph{Richer treatment of structure within special name patterns, dates, etc.} CGEL is largely silent on this point.

\subsection{Potential enhancements}

On the whole, the current granularity of category and function labels has proved workable. We are reluctant to make them finer-grained as this would make it more difficult to create and interpret the main tree structures.

However, additional features or forms of annotation could be added to supplement what is already in the tree. 

\paragraph{Morphology.} While parallel UD parses provide morphological information on tokens, the terminology differs somewhat from that of CGEL.
Morphosyntactic features could be added at the phrase level as well.

\paragraph{Clause types.} Currently, the only explicit subtypes of clause in CGELBank are relative clauses. But CGEL defines an inventory of clause types (e.g., open interrogative, closed interrogative, exclamative) which might be added as supplementary information.

\paragraph{Ellipsis.} \Cref{sec:MiscEllipsis} indicates different kinds of ellipsis, but ellipsis is not made explicit in CGELBank at present.

\paragraph{Constructions beyond surface structure.} Formal frameworks often contain mechanisms for constructions such as passives, raising, and control. 
Such constructions are, of course, addressed in CGEL, but not made explicit in trees. Perhaps they should be added as an additional layer somehow.
The licensors of indirect complements (\cref{sec:compind}) and the distinction between predicative and non-predicative adjuncts could be indicated as well.

\chapter*{Acknowledgements}
We acknowledge Matt Reynolds for helpful comments. We also acknowledge extensive and clarifying discussions with John Payne and Geoff Pullum. We're very grateful for all the support we received in preparing this document.

\bibliographystyle{plainnat}
\bibliography{cgelbank}

\begin{thebibliography}{12}
\providecommand{\natexlab}[1]{#1}
\providecommand{\url}[1]{\texttt{#1}}
\expandafter\ifx\csname urlstyle\endcsname\relax
  \providecommand{\doi}[1]{doi: #1}\else
  \providecommand{\doi}{doi: \begingroup \urlstyle{rm}\Url}\fi

\bibitem[de~Marneffe et~al.(2021)de~Marneffe, Manning, Nivre, and
  Zeman]{de_marneffe-21}
{Marie-Catherine} de~Marneffe, Christopher~D. Manning, Joakim Nivre, and Daniel
  Zeman.
\newblock Universal {D}ependencies.
\newblock \emph{Computational Linguistics}, 47\penalty0 (2):\penalty0 255--308,
  July 2021.
\newblock URL \url{https://doi.org/10.1162/coli_a_00402}.

\bibitem[Huddleston and Pullum(2002)]{cgel}
Rodney Huddleston and Geoffrey~K. Pullum, editors.
\newblock \emph{The {C}ambridge {G}rammar of the {E}nglish {L}anguage}.
\newblock Cambridge University Press, Cambridge, {UK}, 2002.
\newblock URL
  \url{https://archive.org/details/TheCambridgeGrammarOfTheEnglishLanguage}.

\bibitem[Huddleston et~al.(2021)Huddleston, Pullum, and Reynolds]{sieg2}
Rodney Huddleston, Geoffrey~K. Pullum, and Brett Reynolds.
\newblock \emph{A {S}tudent's {I}ntroduction to {E}nglish {G}rammar}.
\newblock Cambridge University Press, 2nd edition, 2021.
\newblock {DOI}: 10.1017/9781009085748.

\bibitem[Levine(2012)]{Levine2012}
Robert~D. Levine.
\newblock Auxiliaries: \textit{To}'s company.
\newblock \emph{Journal of Linguistics}, 48\penalty0 (1):\penalty0 187--203,
  2012.
\newblock \doi{10.1017/S002222671100034X}.
\newblock URL
  \url{http://www.journals.cambridge.org/abstract_S002222671100034X}.

\bibitem[Matthiessen and Bateman(1991)]{matthiessen-91}
Christian M. I.~M. Matthiessen and John~A. Bateman.
\newblock \emph{Text {G}eneration and {S}ystemic-functional {L}inguistics:
  {E}xperiences from {E}nglish and {J}apanese}.
\newblock Pinter, 1991.

\bibitem[Payne et~al.(2007)Payne, Huddleston, and Pullum]{payne-07}
John Payne, Rodney Huddleston, and Geoffrey~K. Pullum.
\newblock Fusion of functions: {T}he syntax of \emph{once}, \emph{twice} and
  \emph{thrice}.
\newblock \emph{Journal of Linguistics}, 43\penalty0 (3):\penalty0 565--603,
  November 2007.
\newblock URL
  \url{https://www.cambridge.org/core/journals/journal-of-linguistics/article/fusion-of-functions-the-syntax-of-once-twice-and-thrice1/D0D583318480CCEB18536C891286D48E}.
\newblock Cambridge University Press.

\bibitem[Payne et~al.(2010)Payne, Huddleston, and Pullum]{payne-10}
John Payne, Rodney Huddleston, and Geoffrey~K. Pullum.
\newblock The distribution and category status of adjectives and adverbs.
\newblock \emph{Word Structure}, 3\penalty0 (1):\penalty0 31--81, April 2010.
\newblock URL
  \url{https://www.euppublishing.com/doi/abs/10.3366/E1750124510000486}.
\newblock Edinburgh University Press.

\bibitem[Payne et~al.(2013)Payne, Pullum, Scholz, and Berlage]{payne-13}
John Payne, Geoffrey~K. Pullum, Barbara~C. Scholz, and Eva Berlage.
\newblock Anaphoric \emph{one} and its implications.
\newblock \emph{Language}, 89\penalty0 (4):\penalty0 794--829, 2013.
\newblock URL \url{https://www.jstor.org/stable/24671958}.
\newblock Linguistic Society of America.

\bibitem[Pullum and Reynolds(2013)]{pullum-13}
Geoffrey~K. Pullum and Brett Reynolds.
\newblock New members of `closed classes' in {E}nglish.
\newblock Manuscript, February 2013.
\newblock URL
  \url{https://www.researchgate.net/publication/260122411_New_members_of_%27closed_classes%27_in_English}.

\bibitem[Pullum and Rogers(2009)]{pullum-09}
Geoffrey~K. Pullum and James Rogers.
\newblock Expressive power of the syntactic theory implicit in \emph{{T}he
  {C}ambridge {G}rammar of the {E}nglish {L}anguage}.
\newblock In \emph{Annual Meeting of the Linguistics Association of Great
  Britain}, pages 1--16, 2009.
\newblock URL \url{http://www.lel.ed.ac.uk/~gpullum/EssexLAGB.pdf}.

\bibitem[Reynolds(2025)]{reynolds2025}
Brett Reynolds.
\newblock Numerical syntax: {T}oward a proper analysis of {E}nglish
  numeratives.
\newblock In preparation, 2025.

\bibitem[Reynolds et~al.(2023)Reynolds, Arora, and Schneider]{cgelbank-law}
Brett Reynolds, Aryaman Arora, and Nathan Schneider.
\newblock Unified syntactic annotation of {E}nglish in the {CGEL} framework.
\newblock In \emph{Proc. of the 17th Linguistic Annotation Workshop
  (LAW-XVII)}, Toronto, Canada, July 2023.
\newblock URL \url{https://doi.org/10.18653/v1/2023.law-1.22}.

\end{thebibliography}

\appendix

\chapter{Justification of lexical coordination analysis}\label{app:lexcoord}

\emph{This appendix offers an in-depth justification of the analysis of lexical coordination presented in \cref{sec:lexcoord}.}

\smallskip

Coordination of lexemes presents the puzzle of what category to assign the expanded coordinate. Strictly, \mention{and will} in \cref{ex:lexemes1,ex:lexemes2,ex:lexemes3,ex:lexemes4,ex:lexemes6a} is neither a verb, a kind of lexeme, since it has internal structure; nor is it a typical VP as it will not admit of any complement. A number of possible analyses present themselves.

\begin{enumerate}
    \item Call it a V and ignore the internal structure in a lexical category as in \cref{ex:lexemes1}.
    \item Forbid coordination of lexemes, and introduce a new set of marked categories which are atypical phrases as in \cref{ex:lexemes2}.
    \item Call all non-marked single coordinates Vs and call marked coordinates VPs, as in \cref{ex:lexemes3}, accepting that such a VP will not allow a complement. (This fact can actually be motivated by the preference for binary branching, noting that a VP like \mention{easily do it} also does not allow a complement.) 
    \item\label{item:chosen} \manicule{} Treat auxiliaries in coordination as projecting unary VPs as the coordinates \cref{ex:lexemes6a}.
    \item Forbid coordination of lexemes, call it a VP, and include a gap co-indexed to the post-nuclear complement as in \cref{ex:lexemes4}. This makes it a delayed right constituent coordination (see \cref{sec:delayed}).
\end{enumerate}

The last option seems appealing for simple cases like \cref{ex:lexemes4} and \cref{ex:lemmasQ}, but for cases like \cref{ex:lemmasQ2}, it would require the head of the main clause to be co-indexed to a prenucleus (because of Subj-Aux inversion), to have that prenucleus have coordinated phrases with gaps and a post-nucleus, but the post-nucleus would not be realized there. Instead it would be a gap co-indexed to a post-nucleus in the main VP. This seems beyond the pale.

On balance, we feel option~\ref{item:chosen} (unary VP coordinates) offers the best compromise. It preserves coordination of like categories, does not require a new category for marked coordinates, and does not apply a lexical category to marked coordinates, which are phrases. Finally, it adheres to the principle that all lexemes other than subordinators and coordinators project a phrasal category (\cref{sec:lexprojection}).

\begin{figure}
\begin{multicols}{2}

\needspace{10em}
\ex.\label{ex:lexemes1} 
V\textsubscript{aux} marked coordinate

    \begin{forest}
    where n children=0{
        font=\itshape, 			
        tier=word          			
     }{
     },
    [Clause
    	[\Subj{NP}[I, roof]]
    	[\Head{VP}
    		[\Head{Coordination}
    			[\Node{Coordinate}{\fbox{V\textsubscript{aux}}}[can]]	[\Node{Coordinate}{\fbox{V\textsubscript{aux}}}
    				[\Mk{Coordinator}[and]]	[\Head{\fbox{V\textsubscript{aux}}}[will]]
    			]
    		]
    		[\Comp{Clause}[try, roof]]
    	]
    ]
    \end{forest}

\needspace{10em}
\ex.\label{ex:lexemes2}
V\textsubscript{aux}\textsuperscript{mk}

    \begin{forest}
    where n children=0{
        font=\itshape, 			
        tier=word          			
     }{
     },
    [Clause
    	[\Subj{NP}[I, roof]]
    	[\Head{VP}
    		[\Head{Coordination}
    			[\Node{Coordinate}{\fbox{V\textsubscript{aux}}}[can]]	[\Node{Coordinate}{\fbox{V\textsubscript{aux}\textsuperscript{mk}}}
    				[\Mk{Coordinator}[and]]	[\Head{\fbox{V\textsubscript{aux}}}[will]]
    			]
    		]
    		[\Comp{Clause}[try, roof]]
    	]
    ]
    \end{forest}

\end{multicols}

\needspace{10em}
\begin{multicols}{2}

\ex.\label{ex:lexemes3}
VP marked coordinate

    \begin{forest}
    where n children=0{
        font=\itshape, 			
        tier=word          			
     }{
     },
    [Clause
    	[\Subj{NP}[I, roof]]
    	[\Head{VP}
    		[\Head{Coordination}
    			[\Node{Coordinate}{\fbox{V\textsubscript{aux}}}[can]]
    			[\Node{Coordinate}{\fbox{VP}}
    				[\Mk{Coordinator}[and]]	[\Head{\fbox{V\textsubscript{aux}}}[will]]
    			]
    		]
    		[\Comp{Clause}[try, roof]]
    	]
    ]
    \end{forest}

\needspace{10em}
\ex.\label{ex:lexemes6a}
Unary VP coordinates

    \begin{forest}
    where n children=0{
        font=\itshape, 			
        tier=word          			
      }{
      },
    [VP
		[\Head{Coordination}
			[\Node{Coordinate}{\fbox{VP}}
			    [\Head{V\textsubscript{aux}}[can]]]
			[\Node{Coordinate}{\fbox{VP}}
				[\Mk{Coordinator}[and]]	[\Head{\fbox{VP}}
				    [\Head{V\textsubscript{aux}}[will]]]
			]
		]
		[\Comp{Clause}[try, roof]]
	]
    \end{forest}

\end{multicols}
    \label{fig:lexcoord1}
\end{figure}

\needspace{10em}
\ex.\label{ex:lexemes4}
Gapped VP coordinates

    \begin{forest}
    where n children=0{
        font=\itshape, 			
        tier=word          			
     }{
     },
    [Clause
    	[\Subj{NP}[I]]
    	[\Head{VP}
    		[\Head{Coordination}
    			[\Node{Coordinate}{\fbox{VP}}
    			    [\Head{V\textsubscript{aux}}[can]]
    			    [\Comp{GAP\textsubscript{x}}[--]]
    			]
    	        [\Node{Coordinate}{\fbox{VP}}
    			    [\Mk{Coordinator}[and]]
        			[\Head{\fbox{VP}}
        			    [\Head{V\textsubscript{aux}}[will]]
        			    [\Comp{GAP\textsubscript{x}}[--]]
        			]
        		]
    		]
    		[\Node{Postnucleus}{Clause\textsubscript{x}}[try, roof]]
    	]
    ]
    \end{forest}

\ex. \label{ex:lemmasQ}
\begin{forest}
where n children=0{
    font=\itshape, 			
    tier=word          			
 }{
 },
[Clause
	[\Node{Prenucleus}{NP\fbox{\textsubscript{x}}}[what, roof]]
	[\Head{Clause}
		[\Node{Prenucleus}{V\textsubscript{aux}\fbox{\textsubscript{y}}}[are]]
		[\Head{Clause}
			[\Subj{NP}[you, roof]]
			[\Head{VP}
			    [\Head{GAP\fbox{\textsubscript{y}}}[--]]
                [\Comp{Clause}
        			[\Head{VP}
        				[\Head{Coordination}
        					[\Node{Coordinate}{VP}
        						[\Head{V}[eating]]
        						[\Obj{GAP\fbox{\textsubscript{x}}}[--]]
        					]
        					[\Node{Coordinate}{VP}
        						[\Mk{Coordinator}[and]]
        						[\Head{VP}
        							[\Head{V}[drinking]]
        							[\Obj{GAP\fbox{\textsubscript{x}}}[--]]
        						]
        					]
        				]
        				[\Node{Postnucleus}{NP\fbox{\textsubscript{x}}}[--]]
        			]
        		]
            ]
		]
	]
]
\end{forest}

\ex. \label{ex:lemmasQ2}
    \begin{forest}
    where n children=0{
        font=\itshape, 			
        tier=word          			
     }{
     },
    [Clause
    	[\Node{Prenucleus}{NP\textsubscript{x}}[what, roof]]
    	[\Head{Clause}
    		[\Node{Prenucleus}{VP\textsubscript{z}}
    			[\Head{Coordination}
    				[\Node{Coordinate}{VP}
    					[\Head{V\textsubscript{aux}}[did]]
    					[\Comp{GAP\textsubscript{y}}[--]]
    				]
    				[\Node{Coordinate}{VP}
    					[\Mk{Coordinator}[or]]
    					[\Head{VP}
    						[\Head{V\textsubscript{aux}}[will]]
    						[\Comp{GAP\textsubscript{y}}[--]]
    					]
    				]
    			]
    			[\Node{Postnucleus}{NP\textsubscript{y}}[--]]
    		]
    		[\Head{Clause}
    			[\Subj{NP}[you, roof]]
    			[\Head{VP}
    			    [\Head{GAP\textsubscript{z}}[--]]
                    [\Node{Postnucleus}{Clause\textsubscript{y}}
                        [\Head{VP}
                            [\Head{V}[decide]]
                            [\Obj{GAP\textsubscript{x}}[--]]
                        ]
                    ]
                ]
    		]
    	]
    ]
    \end{forest}
    
\chapter{Data Format}\label{ch:format}

The data release can be accessed at \url{https://github.com/nert-nlp/cgel}.
The raw data is stored in .cgel files, one per subcorpus: currently, ewt.cgel for the EWT trees and twitter.cgel for the Twitter trees.
These files can be opened in a text editor.
A Python API for loading the trees is provided for scripting (see cgel.py), and there is also a tool (tree2tex.py) to generate LaTeX to produce graphical versions of the trees.

An example raw tree is given in \cref{fig:rawtree}. 
Each tree has two parts: the header\slash metadata section 
consisting of lines beginning with \texttt{\#}, 
and the tree itself.

Lines in the header are key-value pairs following the .conllu standard used in the Universal Dependencies project. 
In particular, every sentence has an ID as well as a number indicating its order within the file. 
The \texttt{text} line indicates the original, untokenized sentence. 
The \texttt{sent} line is the sequence of terminals to appear in the CGEL tree: punctuation tokens are removed, capitalization is normalized, and \verb|--| is inserted at positions where there are gaps.

The tree itself is in a parenthesized format adapted from PENMAN notation \citep{matthiessen-91}. 
Every line represents the start of a constituent.
Parentheses (and accompanying indentation) indicate the bracketing structure. 
Functions begin with a \texttt{:} symbol and are capitalized. 
Gaps in the tree consist of the GAP category.
A coindexation variable and slash precede gaps and their coindexed constituents.

\begin{figure}
    \centering\small
    \begin{verbatim}
# sent_id = Tree IsThatWhatYouCall-0
# sent_num = 4
# text = Is that what you call WH-movement?
# sent = is that -- what you call -- WH-movement
(Clause
    :Prenucleus (x / VP
        :Head (V_aux :t "is" :l "be" :xpos "VBZ"))
    :Head (Clause
        :Subj (NP
            :Head (Nom
                :Det-Head (DP
                    :Head (D :t "that"))))
        :Head (VP
            :Head (x / GAP)
            :PredComp (NP
                :Head (Nom
                    :Mod (Clause_rel
                        :Head-Prenucleus (y / NP
                            :Head (Nom
                                :Head (N_pro :t "what")))
                        :Head (Clause_rel
                            :Subj (NP
                                :Head (Nom
                                    :Head (N_pro :t "you")))
                            :Head (VP
                                :Head (V :t "call" :xpos "VBP")
                                :Obj_dir (y / GAP)
                                :Obj_ind (NP
                                    :Head (Nom
                                        :Head (N :t "WH-movement" 
                                            :subt "WH" :subt "-" 
                                            :subt "movement" :p "?")))))))))))
    \end{verbatim}
    \caption{Example raw tree from twitter.cgel (with extra line breaks in the final \mention{WH-movement} constituent so it doesn't overflow the margin). The graphical view of this tree is in \cref{fig:graphtree}.}
    \label{fig:rawtree}
\end{figure}

\newcommand{\idx}[1]{\textsubscript{#1}}

\begin{figure}
    \centering
    \begin{forest}
        where n children=0{
            font=\itshape, 			
            tier=word          			
          }{
          },
        [Clause
    [\Node{Prenucleus}{V\textsubscript{aux}\idx{x}}[is]]
    [\Node{Head}{Clause}
        [\Node{Subj}{NP}
            [X,phantom]
            [\Node{Head}{Nom}, before drawing tree={x+=1.5em}
                [\Node{Det-Head}{DP}, no edge
                    [\Node{Head}{D}[that]]] { \draw[-] (!uu.south) -- (); \draw[-] (!u.south) -- (); }]
            [Y,phantom]]
        [\Node{Head}{VP}
            [\Node{Head}{GAP\idx{x}}[--]]
            [\Node{PredComp}{NP}
                [\Node{Head}{Nom}
                    [\Node{Mod}{Clause\textsubscript{rel}}, before drawing tree={x+=4em}
                        [\Node{Head-Prenucleus}{NP\idx{y}}, no edge
                            [\Node{Head}{Nom}
                                [\Node{Head}{N\textsubscript{pro}}[what]]]] { \draw[-] (!uu.south) -- (); \draw[-] (!u.south) -- (); }
                        [\Node{Head}{Clause\textsubscript{rel}}
                            [\Node{Subj}{NP}
                                [\Node{Head}{Nom}
                                    [\Node{Head}{N\textsubscript{pro}}[you]]]]
                            [\Node{Head}{VP}
                                [\Node{Head}{V}[call]]
                                [\Node{Obj\textsubscript{dir}}{GAP\idx{y}}[--]]
                                [\Node{Obj\textsubscript{ind}}{NP}
                                    [\Node{Head}{Nom}
                                        [\Node{Head}{N}[WH-movement]]]]]]]]]]]]
    \end{forest}
    \caption{Graphical view of the tree from \cref{fig:rawtree}.}
    \label{fig:graphtree}
\end{figure}

\section{Features}

\begin{table}
    \centering\small
    \begin{tabular}{lp{33em}}
        \texttt{:note} & a comment on the analysis \\
        \texttt{:p} & punctuation token before or after a word token (may be used multiple times in the node; see \cref{sec:punc-enc}) \\
        \texttt{:t} & token value: the original form of the word after tokenization (leaf nodes only; GAPs and words inserted to correct an omission have no \texttt{:t}) \\
        \texttt{:subt} & subtoken (used multiple times per node) for words where the Universal Dependencies tokenization is finer-grained, e.g.~possessive clitics \\
        \texttt{:correct} & corrected form (see \cref{sec:errors}) \\
        \texttt{:l} & lemma \emph{when distinct from the word form} \\
        \texttt{:xpos} & morphological category for verbs and numbers (see \cref{sec:xpos})
    \end{tabular}
    \caption{String-valued features that may be specified on nodes in the .cgel format.}
    \label{tab:feats}
\end{table}

\Cref{tab:feats} lists the features that may be indicated within a node following its category. 
These principally apply to lexical nodes, but a \texttt{:note} can appear on any node.
String values for these features must be delimited by double quotes. Two escapes are provided: \verb|\"| for the literal quotation mark character and \verb|\\| for the backslash character.

\section{Fusion}

Fusion of functions, which occurs in 2~places in \cref{fig:graphtree}, is not expressed directly in the raw tree format. The parenthesized notation pretends that the first of the two incoming branches (the one skipping a level) is missing, as in \cref{fig:rawtree}. The branch can be recovered automatically by attaching each constituent in \func{Det-Head}, \func{Mod-Head}, \func{Marker-Head}, or \func{Head-Prenucleus} function to its non-immediate ancestor (typically grandparent) of the appropriate category (this extra attachment reflects the first part of the hyphenated function). For example, the deeper parent of a constituent labeled \func{Det-Head} will be a Nom, and its additional parent will be the NP headed by that Nom (or, if there is layering as in \cref{ex:fused-2postmods}, another Nom which it heads).

\end{document}